# SRRM: A Retrieval-Reconstruction Benchmark for Evaluating Long-Context Memory in Small Language Models

**Christopher Alberto Hamonangan**
SMAN 8 Jakarta
Bukit Duri, South Jakarta 12840
calbertohamonangan@gmail.com

## Abstract

Small Language Models (SLMs) are increasingly deployed on consumer-grade hardware, yet their ability to preserve information across progressively longer contexts remains insufficiently understood. Existing benchmarks primarily evaluate retrieval-oriented tasks, providing limited insight into whether successful retrieval reflects comprehensive long-context memory preservation. To address this limitation, this paper introduces Simple Retrieval–Reconstruction Memory (SRRM), a lightweight and fully reproducible benchmark that jointly evaluates Targeted Retrieval and Global Information Reconstruction. This study also proposes the Memory Preservation Ratio (MPR), a benchmark-level metric that quantifies the relationship between retrieval and reconstruction performance. SRRM was evaluated using a synthetic dataset on three Qwen2.5 models (0.5B, 1.5B, and 3B parameters) across context loads ranging from 10 to 100 factual statements. Experimental results show that reconstruction performance deteriorated substantially faster than retrieval accuracy as contextual information increased, indicating that retrieval alone is insufficient to characterize long-context memory preservation. Among the evaluated models, Qwen2.5-1.5B achieved the highest retrieval accuracy (**86.7%**), reconstruction performance (**45.7%**), and MPR (**0.53**), outperforming the larger Qwen2.5-3B model (**72.2%, 18.5%, and 0.26, respectively**). These findings demonstrate that increasing parameter count does not necessarily improve long-context memory preservation. Overall, SRRM provides a reproducible and extensible benchmark for systematically evaluating long-context memory while revealing complementary memory behaviors that remain hidden under retrieval-only evaluation frameworks.

## 1 Introduction

Large Language Models (LLMs) have demonstrated remarkable performance across a wide range of tasks, including conversational assistance, code generation, reasoning, and document understanding [13, 19]. While frontier proprietary models continue to achieve state-of-the-art capabilities [13], their deployment typically relies on large-scale cloud infrastructures that are inaccessible for offline or resource-constrained environments. Consequently, recent years have witnessed the rapid development of Small Language Models (SLMs), which significantly reduce model size while maintaining competitive performance for many practical applications. Their relatively low computational requirements make them increasingly attractive for deployment on consumer-grade hardware such as laptops, edge devices, and smartphones, often utilizing efficient parameter adaptation techniques [7] or modular extensions [9].

Despite these advantages, the practical use of SLMs introduces a unique challenge. Although many modern models advertise context windows of up to 128K tokens or more, resource limitations on local devices often force users to operate them with substantially smaller context sizes. Under such constraints, users must carefully balance the amount of contextual information supplied to the model against computational efficiency and response quality, as models inherently struggle to utilize long contexts effectively without loss of focus [15, 12, 16]. Understanding how information retention degrades as the amount of contextual information increases therefore becomes an important problem for both researchers and practitioners, especially in dynamic, multi-turn application scenarios [8, 13].

Existing evaluation benchmarks primarily focus on downstream task performance, reasoning ability, long-context question answering [15], or targeted information retrieval [13, 17]. However, relatively few studies systematically investigate how information retention changes as contextual information progressively increases [11, 18], particularly for small language models deployed in local environments. Furthermore, existing benchmarks generally evaluate a single retrieval-oriented task [15], leaving unanswered whether the ability to retrieve an individual fact is accompanied by an equivalent ability to reconstruct an entire collection of previously presented information, given that factual predictions are highly sensitive to local context distortions [18] and long-tail knowledge rarity [10].

To address this gap, this paper introduces Simple Retrieval–Reconstruction Memory (SRRM), a lightweight and fully reproducible benchmark designed to evaluate information retention from two complementary perspectives. The first evaluates Targeted Retrieval, which measures a model's ability to recall a specific fact from previously presented information [13, 17]. The second evaluates Global Information Reconstruction, which measures how completely the model can reconstruct all previously presented facts without being prompted for individual items. By jointly evaluating these two tasks, SRRM provides a more comprehensive characterization of information retention behavior than retrieval-only evaluations.

Using a synthetic dataset generated under controlled conditions, SRRM evaluates three members of the Qwen2.5 model family (0.5B, 1.5B, and 3B parameters) across progressively increasing context loads containing 10, 25, 50, 75, and 100 factual statements. Besides retrieval accuracy and reconstruction performance, the benchmark also records inference latency and token consumption, enabling a joint analysis of memory retention and computational efficiency. The experimental results provide new insights into how model scale influences information retention under realistic local deployment scenarios [1, 10, 15].

Based on the identified research gap, this study seeks to answer the following research questions:
**RQ1:** How does retrieval performance change as contextual information progressively increases in small language models?

**RQ2:** Does successful targeted retrieval necessarily correspond to equally effective global information reconstruction under increasing context loads?

**RQ3:** How does model scale influence long-context memory preservation when evaluated using complementary retrieval and reconstruction tasks?

To address these research questions, this paper makes the following contributions:

- **It introduces Simple Retrieval-Reconstruction Memory (SRRM)**, a lightweight and fully reproducible benchmark for evaluating long-context memory in small language models.

- **It proposes a complementary evaluation framework that jointly measures Targeted Retrieval and Global Information Reconstruction**, providing a broader characterization of long-context memory than retrieval-only benchmarks.

- **It introduces the Memory Preservation Ratio (MPR)**, a benchmark-level metric that quantifies the relationship between retrieval performance and reconstruction capability, enabling a more integrated assessment of information preservation.

- **It provides the first systematic SRRM evaluation of the Qwen2.5 model family across progressively increasing context loads**, revealing benchmark behaviors that are not immediately apparent from retrieval performance alone.

- **It releases the complete benchmark implementation**, synthetic dataset generator, evaluation pipeline, and experimental configuration to support reproducibility and facilitate future benchmark extensions by the research community.

To promote reproducible research, the complete implementation of SRRM—including the benchmark dataset generator, evaluation pipeline, source code, and experimental configuration—will be publicly released through an open-source repository upon publication. This repository is intended to facilitate independent verification, comparative benchmarking, and future extensions of the proposed framework.

## 2 Related Work

The rapid advancement of Large Language Models (LLMs) has been accompanied by substantial progress in long-context modeling. Recent models are capable of processing context windows ranging from tens of thousands to more than one hundred thousand tokens, enabling applications involving document understanding, multi-document reasoning, conversational memory, and code analysis. However, increasing the maximum context window does not necessarily imply that models can effectively utilize or retain all information contained within long contexts. Consequently, a growing number of benchmarks have been proposed to evaluate the practical memory capabilities of modern language models under controlled experimental settings.

Table 1 : Comparison of Existing Long-Context Benchmarks

| Benchmark | Primary Objective | Evaluation Domain | Main Task |
|---|---|---|---|
| Lost in the Middle | Positional information retrieval | Text | Multi-document QA, Key-Value Retrieval |
| MMNeedle | Visual information retrieval | Multimodal | Needle localization |
| LongBench | Long-context understanding | Text | Multiple downstream tasks |
| **SRRM** | Information retention | Text | Retrieval and Reconstruction |

Among the earliest and most influential studies, Lost in the Middle proposed by Liu et al. (2023) challenged the common assumption that larger context windows inherently lead to better information utilization. The authors introduced several controlled evaluation protocols to analyze information retention in Transformer-based architectures. Their primary evaluation employed Multi-Document Question Answering, where a target document containing the correct answer was intentionally placed at different positions among numerous distractor documents. In addition, the study proposed a purely synthetic benchmark called Synthetic Key-Value Retrieval, in which randomly generated UUID-based key-value pairs eliminated semantic biases and isolated the model's ability to retrieve factual information from long token sequences. The paper further extended its evaluation using Retriever-Reader pipelines for Open-Domain Question Answering to investigate the trade-off between increasing contextual information and downstream task performance. The results demonstrated that retrieval performance strongly depends on the positional placement of relevant information, revealing that information located near the beginning or the end of the context is significantly easier to retrieve than information positioned near the middle.

Building upon textual long-context evaluation, Wang et al. (2025) introduced MultiModal Needle-in-a-Haystack (MMNeedle) to extend long-context benchmarking into multimodal settings. Instead of textual retrieval, MMNeedle evaluates whether Multimodal Large Language Models (MLLMs) can successfully identify a target visual object hidden among numerous distracting images based solely on textual instructions. The benchmark systematically increases visual context complexity through two independent dimensions, namely the number of input images and the visual information density generated through image stitching. Evaluation is performed using hierarchical metrics consisting of Existence Accuracy, Index Accuracy, and Exact Accuracy, together with negative samples in which the target object is intentionally absent. Experimental results revealed a substantial performance gap between proprietary API-based models and open-source MLLMs, while also demonstrating that increasing visual context complexity significantly degrades retrieval performance even for the strongest commercial models.

A broader evaluation framework was introduced through LongBench by Bai et al. (2024). Rather than focusing exclusively on retrieval tasks, LongBench provides the first bilingual benchmark covering both English and Chinese across twenty-one datasets and six categories of real-world applications, including question answering, summarization, code completion, few-shot learning, and synthetic reasoning tasks. By incorporating documents containing several thousand words, LongBench enables comprehensive evaluation of long-context understanding across diverse downstream scenarios. Unlike synthetic retrieval benchmarks that isolate a single memory behavior, LongBench emphasizes practical task performance under naturally occurring long contexts.

**Table 2 :** Comparison of the memory-related capabilities evaluated by representative long-context benchmarks. A check mark (✓) indicates that the corresponding capability is explicitly evaluated, a cross (✗) indicates that it is not included, and a partial, indicates that the capability is only indirectly or partially assessed

**Table 3:** Comparison of experimental characteristics across representative benchmarks. Unlike previous benchmarks, SRRM is designed for reproducible evaluation of locally deployed Small Language Models while simultaneously measuring memory retention and computational efficiency.

| Capability | LitM | MMN | LB | SRRM |
|---|---|---|---|---|
| Target Retrieval | ✓ | ✓ | ✓ | ✓ |
| Reconstruction | ✗ | ✗ | ✗ | ✓ |
| Progressive Context | Partial | Partial | Partial | ✓ |
| Synthetic Data | ✓ | ✓ | ✓ | ✓ |

| Characteristic | LitM | MMN | LB | SRRM |
|---|---|---|---|---|
| Local SLM Focus | ✗ | ✗ | ✗ | ✓ |
| Deterministic Protocol | ✓ | ✓ | ✓ | ✓ |
| Fully Reproducible | ✓ | ✓ | ✓ | ✓ |
| Computational Metrics | ✗ | ✗ | Partial | ✓ |

Collectively, these benchmarks have significantly advanced the evaluation of long-context capabilities by examining retrieval performance, reasoning ability, multimodal understanding, and downstream applications. Nevertheless, most existing benchmarks primarily assess whether a model can retrieve one specific target item or successfully complete a downstream task. Comparatively less attention has been devoted to understanding whether models can simultaneously retain and reconstruct an entire collection of previously presented factual information as contextual load progressively increases. Consequently, retrieval-oriented evaluations alone may not fully characterize the practical memory behavior of language models deployed under realistic local computing constraints.

Motivated by these observations, the proposed Simple Retrieval-Reconstruction Memory (SRRM) benchmark complements existing long-context benchmarks by jointly evaluating two fundamental aspects of information retention: Targeted Retrieval, which measures the ability to recall individual factual statements, and Global Information Reconstruction, which measures the ability to reconstruct the complete set of previously observed information without explicit guidance. Rather than evaluating reasoning performance or downstream task completion, SRRM focuses specifically on memory retention under progressively increasing factual contexts while maintaining a lightweight, fully reproducible experimental design suitable for locally deployed Small Language Models.

**Table 4: Comparison between existing benchmark objectives and the proposed SRRM benchmark.** SRRM extends previous evaluations by jointly measuring targeted retrieval, global information reconstruction, and computational efficiency under progressively increasing factual contexts.

| Aspect | Existing Benchmarks | SRRM |
|---|---|---|
| Evaluation objective | Retrieval or downstream performance | Information retention |
| Memory granularity | Individual target facts | Individual and global information |
| Context scaling | Long documents or images | Progressive factual contexts |
| Evaluation outputs | Task-specific metrics | Retrieval Accuracy, Global Information Reconstruction, latency, and token usage |
| Target deployment | General-purpose LLMs | Locally deployed Small Language Models |

# 3 Methodology

This study introduces the Short-term Retrieval and Reconstruction Memory (SRRM) benchmark for evaluating information retention in small language models. To minimize confounding variables, the experiments focus on models from the same family with different parameter scales, enabling a controlled investigation of the relationship between model size and memory performance. Since no existing benchmark satisfies the objectives of this study, a synthetic dataset generator was developed to automatically construct structured factual information with consistent characteristics. This design ensures that all evaluated models receive identical inputs under reproducible experimental conditions.

## 3.1 Research Design

This study adopts a quantitative experimental benchmark design to evaluate short-term information retention in small language models under controlled conditions. Rather than comparing different model families, the experiments focus exclusively on the Qwen2.5 model family, where the primary independent variable is the model parameter scale (0.5B, 1.5B, and 3B). Restricting the evaluation to a single model family minimizes architectural and training-related confounding factors, allowing the benchmark to investigate the relationship between parameter scaling and memory performance more objectively.

For each experiment, the benchmark automatically generates a fixed set of structured factual information using the proposed synthetic dataset generator. The generated dataset remains identical across all evaluated models to ensure fair comparison and eliminate variations caused by data sampling. Each experiment is repeated under five context sizes consisting of 10, 25, 50, 75, and 100 information items.

**Figure 1:** Overall architecture of the proposed SRRM benchmark. Synthetic information is generated and organized into conversational context before being evaluated through Sequential Retrieval and Summary Reconstruction tasks. Performance is measured using retrieval accuracy, Critical Information Recall (CIR), token consumption, inference time, and hardware memory **usage.**

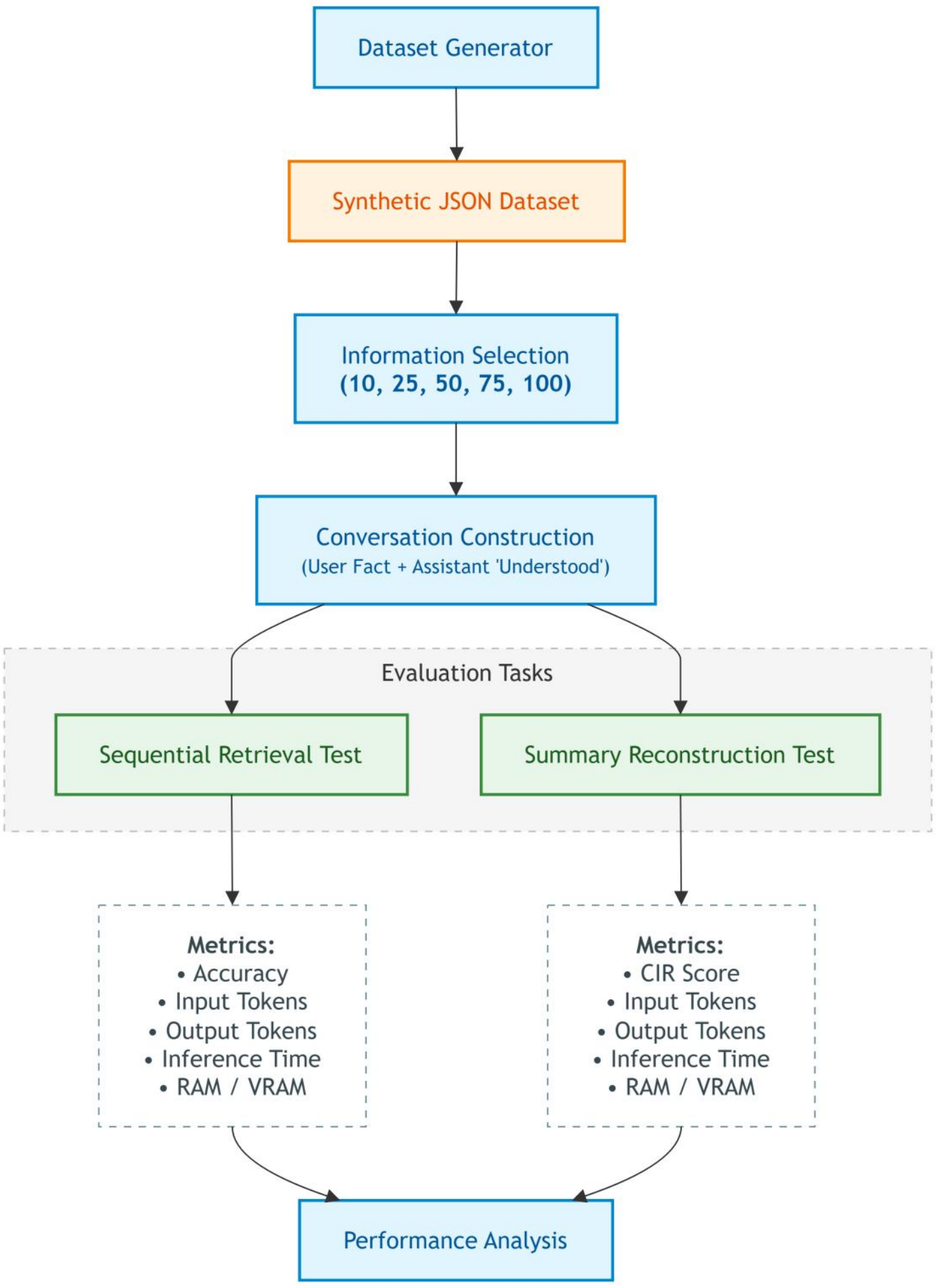

The experimental pipeline consists of two sequential evaluation stages. During the first stage, all generated facts are inserted into a simulated conversation history before the model is evaluated through a Sequential Retrieval Test, where factual questions are asked individually while preserving the same contextual information throughout the evaluation. The second stage performs a Summary Reconstruction Test, requiring the model to reconstruct every previously presented fact from memory without receiving additional cues.

Model performance is evaluated using five quantitative metrics: Retrieval Accuracy, Critical Information Recall (CIR), input-output token consumption, inference time, and memory usage (RAM and VRAM). Together, these measurements provide complementary perspectives on retrieval capability, information reconstruction, computational efficiency, and hardware resource utilization, enabling a comprehensive analysis of information retention under increasing context loads.

## 3.2 Models

To investigate the relationship between model parameter scaling and short-term information retention, this study evaluates three instruction-tuned models from the Qwen2.5 family. Restricting the evaluation to a single model family minimizes architectural, tokenizer, and training-related confounding factors, allowing differences in memory performance to be primarily attributed to model parameter size rather than variations in model design.

All experiments were conducted using identical inference settings to ensure a fair comparison across models. Every model received the same synthetic dataset, identical conversation history, identical prompts, and identical evaluation procedures throughout the benchmark. The evaluated models are summarized in Table 5, while the inference configuration and hardware environment are presented in Table 6 and Table 7, respectively.

**Table 5:** As shown in Table 5, the benchmark evaluates three parameter scales within the Qwen2.5 family while maintaining the same inference framework and context length. This controlled selection enables the study to isolate the effect of parameter scaling on information retention performance.

| Model | Parameters | Quantization | Context Length | Inference Framework |
|---|---|---|---|---|
| Qwen2.5:0.5B | 0.5 Billion | *(Auto / Ollama)* | 4096 | Ollama |
| Qwen2.5:1.5B | 1.5 Billion | *(Auto / Ollama)* | 4096 | Ollama |
| Qwen2.5:3B | 3 Billion | *(Auto / Ollama)* | 4096 | Ollama |

**Table 6:** Inference settings and experimental configuration shared by all evaluated models.

| Configuration | Value |
|---|---|
| Framework | Ollama |
| Temperature | 0.0 |
| Context Length (*num_ctx*) | 4096 |
| Streaming Mode | FALSE |
| Dataset Type | Fixed Synthetic Dataset |
| Context Sizes | 10, 25, 50, 75, 100 Information Items |

**Table 7:** Hardware specifications used for all benchmark experiments..

| Component | Specification |
|---|---|
| Operating System | Windows 11 |
| CPU | AMD Ryzen 7 260 |
| GPU | NVIDIA GeForce 5050 |
| VRAM | 8 GB |
| System RAM | 16 GB |

The standardized model selection, inference configuration, and hardware environment presented in this section establish a controlled experimental setting for the SRRM benchmark. By maintaining identical evaluation conditions across all experiments, the benchmark minimizes external sources of variation and ensures that any observed differences in retrieval or reconstruction performance primarily reflect the characteristics of the evaluated language models. The following section describes the synthetic dataset and its construction process, which forms the foundation of the proposed benchmark.

## 3.3 Dataset

Unlike previous information retention benchmarks that rely on existing public datasets, the proposed SRRM benchmark employs a fully synthetic dataset specifically constructed for this study. The dataset was designed to provide structured factual information while maintaining a controlled and reproducible evaluation environment. By generating every information item programmatically, the benchmark ensures that all evaluated language models receive identical inputs under the same experimental conditions, thereby eliminating variations introduced by random data sampling or dataset inconsistencies.

**Table 8:** This table summarizes the overall characteristics of the synthetic dataset used throughout this study. The dataset is entirely synthetic and stored in a structured JSON format, enabling automatic benchmark generation and objective evaluation. To ensure fairness and reproducibility, each semantic category contains an equal number of generated records, and the complete dataset is kept identical across all experiments. Furthermore, a deterministic Round-Robin organization strategy is adopted to distribute information uniformly within the benchmark sequence, preventing category concentration during evaluation.

| Dataset | Type | Format | Value | Description |
|---|---|---|---|---|
| Dataset | Synthetic | JSON | 160 Information Items | Automatically generated dataset used for all benchmark experiments |
| Categories | Semantic | Structured | 4 | Hair Color, Family Relationship, Activity, and WiFi Password |
| Records | Fixed | Per Category | 40 | Equal number of records generated for each category |
| Language | Natural Language | English | - | All benchmark statements are written in English |
| Distribution | Organization | Round-Robin | - | Information is interleaved evenly across all categories |
| Consistency | Evaluation | Fixed | - | Identical dataset used for every evaluated model |

The synthetic dataset is automatically generated using a custom data generation framework developed specifically for this benchmark. Instead of collecting real-world documents, the generator constructs factual statements from predefined vocabulary pools, including person names, hair colors, food items, geographic locations, and randomly generated WiFi passwords. Each generated statement is accompanied by its corresponding structured ground-truth representation consisting of a subject, an attribute, and an expected value. This design enables both retrieval-based and reconstruction-based evaluations to be performed objectively without manual annotation.

**Table 9:** The synthetic dataset is composed of four semantic information categories designed to represent different forms of factual memory while maintaining a unified subject–attribute–value representation. The Hair Color category evaluates the retention of simple attribute-based facts associated with a single entity. The Family Relationship category introduces relational information between two entities, requiring the model to preserve both the relationship type and the linked individual. The Activity category represents event-based facts by describing either an action performed by a person or a visited location, thereby increasing semantic diversity within the dataset. Finally, the WiFi Password category evaluates the retention of arbitrary alphanumeric strings that contain little semantic meaning, making them significantly more difficult to memorize than ordinary natural-language facts. Despite these semantic differences, every information item is ultimately represented using the same structured subject–attribute–value format, enabling consistent automatic evaluation across all benchmark tasks.

| Category | Statement Pattern | Ground Truth Structure |
|---|---|---|
| Hair Color | *X's hair color is Y.* | Subject, Attribute, Value |
| Family Relationship | *X has a younger/older brother/sister named Y.* | Subject, Attribute, Value |
| Activity | *X ate Y. / X went to Y.* | Subject, Attribute, Value |
| WiFi Password | *X's WiFi password is Y.* | Subject, Attribute, Value |

The four information categories presented in **Table 9** were intentionally selected to introduce varying levels of semantic complexity while preserving a consistent sentence structure. Rather than focusing on a single type of factual knowledge, the dataset combines attribute-based, relational, event-based, and arbitrary symbolic information to provide a broader evaluation of short-term memory behavior. This diversity enables the proposed SRRM benchmark to assess not only whether a language model can retain factual information, but also whether different semantic characteristics influence retrieval and reconstruction performance under increasing context loads.

**Figure 2:** Overall organization of the proposed SRRM benchmark. The benchmark begins with four independently generated synthetic information categories, which are subsequently organized into a balanced benchmark sequence using a deterministic Round-Robin strategy. Each information item is assigned a sequence identifier and category identifier before being grouped into predefined context sizes. The constructed benchmark is then evaluated through two complementary tasks, namely Sequential Retrieval and Summary Reconstruction, while multiple performance metrics are collected to provide a comprehensive analysis of the short-term memory capability of the evaluated language models.

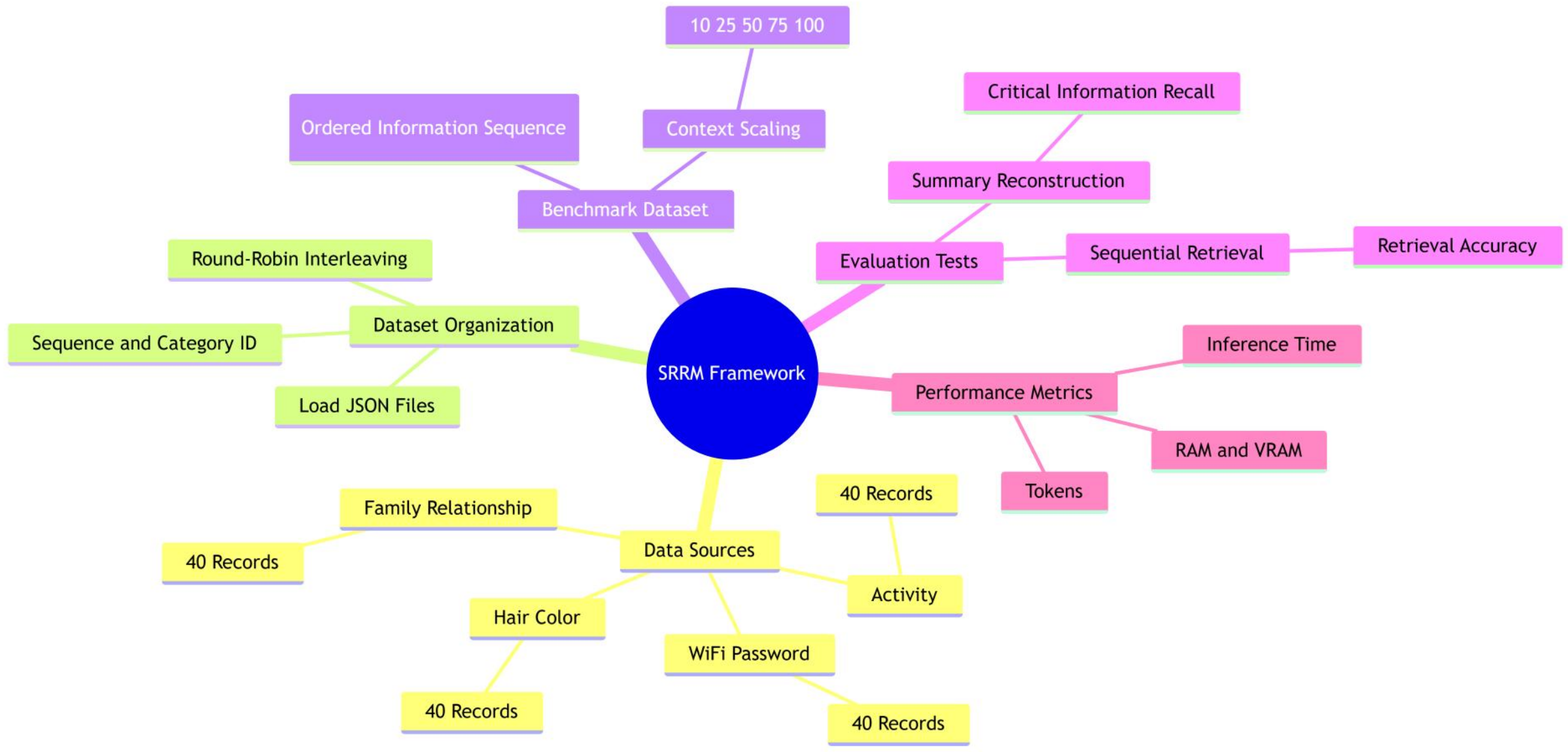

As illustrated in Figure 2, the proposed SRRM benchmark follows a structured evaluation pipeline consisting of four major stages: dataset preparation, benchmark organization, evaluation, and performance analysis. The benchmark first constructs a balanced sequence of synthetic factual information from multiple semantic categories before progressively increasing the context size during evaluation. The organized benchmark is subsequently assessed using both retrieval-based and reconstruction-based tasks, allowing the study to measure not only whether individual facts can be recalled accurately, but also how well the overall information content can be reconstructed. Finally, quantitative performance metrics are collected to characterize the memory retention behavior of each evaluated language model under identical experimental conditions.

After the synthetic dataset is generated, the benchmark organizes all information items through a deterministic Round-Robin interleaving strategy. Instead of presenting information from a single category consecutively, statements from the four categories are inserted alternately to produce a balanced sequence throughout the benchmark. During this process, each information item is assigned both a **sequence identifier**, representing its chronological position in the benchmark, and a **category identifier**, indicating its semantic category. The resulting ordered dataset serves as the input for all subsequent evaluation procedures.

**Figure 3 :** This figure illustrates the organization process performed after the synthetic JSON datasets have been generated. The four semantic categories are first loaded independently before being merged through a deterministic Round-Robin interleaving strategy, ensuring that information from different categories is distributed uniformly throughout the benchmark sequence. During this organization process, each information item is assigned a sequence identifier and a category identifier, producing a standardized benchmark dataset that can subsequently be partitioned into different context sizes for evaluation.

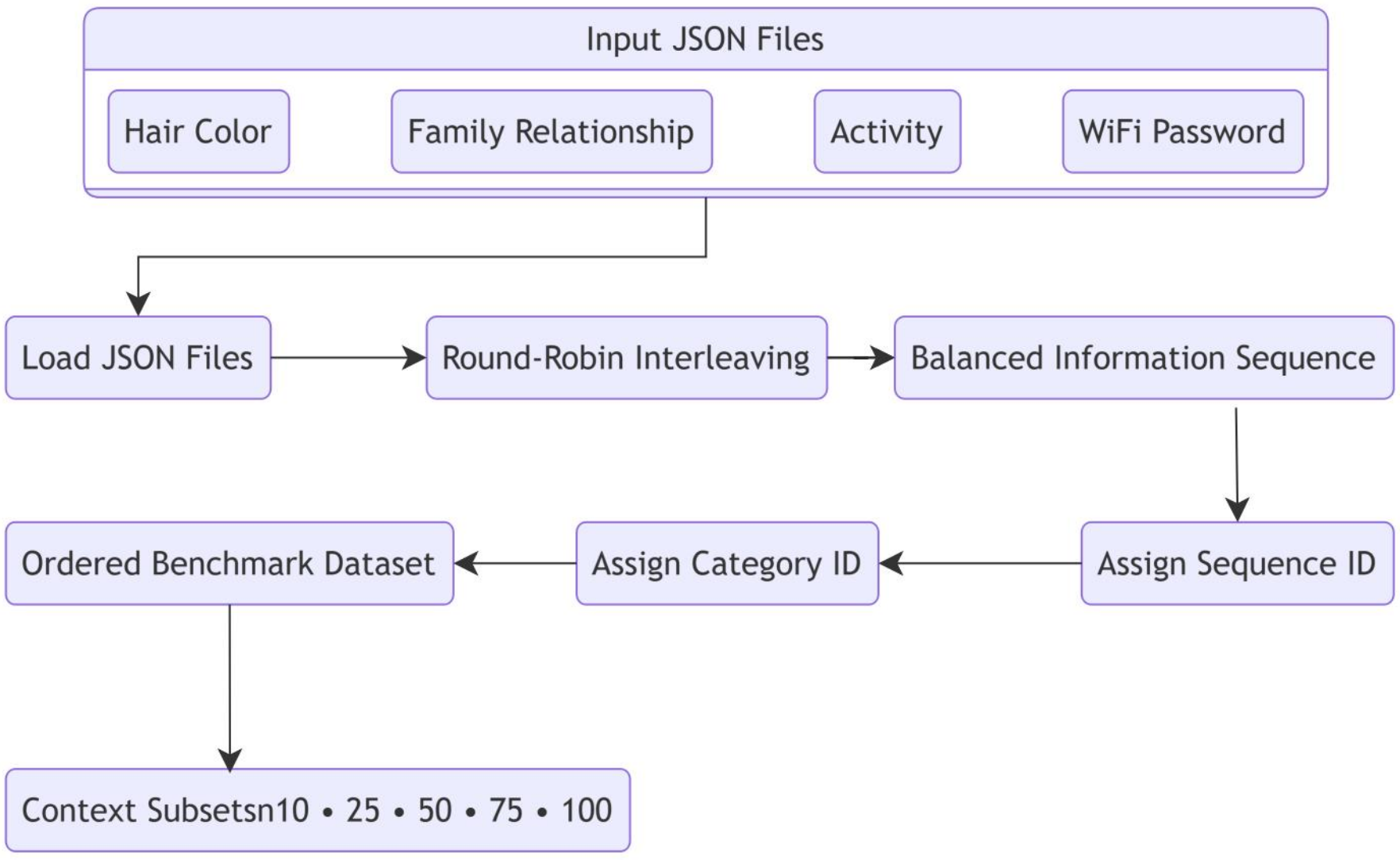

**Table 10:** Each benchmark information record represents a single factual statement presented to the evaluated language model. Besides the natural-language statement itself, every record contains metadata describing its chronological position within the benchmark and its semantic category. These metadata enable deterministic dataset organization, reproducible context construction, and consistent evaluation across different benchmark configurations.

| Field | Description |
|---|---|
| sequence_id | Chronological order of the information within the benchmark |
| category_id | Identifier of the semantic information category |
| text | Natural-language statement presented to the language model |

**Table 11:** Each benchmark record includes a structured ground-truth annotation for automatic evaluation. Instead of relying on raw text comparison, every fact is represented using a standardized subject-attribute-value schema. This representation enables objective verification for both the Sequential Retrieval and Summary Reconstruction tests while minimizing ambiguity during automatic scoring.

| Field | Description |
|---|---|
| ground_truth.subject | Target entity contained in the statement |
| ground_truth.attribute | Semantic relation associated with the entity |
| ground_truth.value | Expected answer used for evaluation |

The dataset organization process illustrated in Figure 3 ultimately produces benchmark records following the structure presented in Table 10. Each organized record consists of both descriptive metadata and the natural-language statement that is delivered to the evaluated language model during benchmark execution. To enable fully automatic and reproducible evaluation, every benchmark record is further paired with the structured ground-truth representation shown in Table 11. Together, these two complementary data structures establish the foundation of the SRRM benchmark by separating the information presented to the model from the reference annotations used during performance assessment.

To evaluate memory retention under increasing context loads, the benchmark performs experiments using five predefined context sizes. Rather than modifying the dataset itself, the benchmark progressively increases the number of information items presented to the model while preserving their original ordering. This design enables the benchmark to analyze how retrieval and reconstruction performance change as the amount of stored information increases.

**Table 11:** The SRRM benchmark evaluates language models under five progressively increasing context sizes to analyze the effect of information load on short-term memory performance. Instead of modifying the dataset itself, the benchmark gradually increases the number of information items presented to the model while preserving the same information order and distribution. This controlled context scaling enables consistent comparisons across different parameter sizes under identical experimental conditions.

| Evaluation Level | Number of Information Items |
|---|---|
| Level 1 | 10 |
| Level 2 | 25 |
| Level 3 | 50 |
| Level 4 | 75 |
| Level 5 | 100 |

The synthetic dataset described in this section forms the foundation of the proposed SRRM benchmark. By combining automated dataset generation, deterministic Round-Robin organization, standardized ground-truth annotations, and fixed context scaling, the benchmark provides a controlled and fully reproducible evaluation environment. These design choices ensure that every evaluated language model is exposed to identical information under the same experimental conditions, allowing subsequent analyses to attribute observed performance differences primarily to model capability rather than variations in the evaluation data.

## 3.4 Evaluation Metrics

The proposed SRRM benchmark evaluates short-term memory performance from both **retrieval** and **reconstruction** perspectives to provide a comprehensive assessment of information retention in language models. Rather than relying on a single evaluation criterion, the benchmark measures the ability to recall individual facts accurately through the Sequential Retrieval Test while simultaneously assessing the capability to reconstruct previously presented information using the Summary Reconstruction Test. These complementary evaluation tasks capture different aspects of memory behavior under progressively increasing context sizes.

In addition to memory performance, the benchmark records several computational metrics throughout the evaluation process, including token consumption, inference time, and hardware memory utilization. Collectively, these measurements provide both effectiveness and efficiency perspectives, enabling a more comprehensive comparison of language models beyond retrieval accuracy alone.

### 3.4.1 Sequential Retrieval Test

The Sequential Retrieval Test (SRT) evaluates a language model's ability to accurately retrieve previously presented factual information from the constructed conversation history. After all information statements have been sequentially inserted into the conversation context, the benchmark generates one retrieval question for every information item. Each question targets a single ground-truth fact, allowing every stored information item to be evaluated independently.

The retrieval questions are evaluated sequentially in reverse chronological order, beginning with the most recently presented information and continuing until the earliest information item. This evaluation strategy enables the benchmark to analyze memory retention across the entire context while maintaining a deterministic and reproducible testing procedure. Since every retrieval question corresponds to exactly one benchmark record, each information item contributes equally to the final retrieval score.

**Figure 4:** This figure illustrates the evaluation workflow of the Sequential Retrieval Test. After the complete conversation history has been constructed, benchmark questions are automatically generated from the corresponding ground-truth annotations. Each model response is subsequently compared with the expected answer to determine whether the retrieved fact is correct or incorrect. The final retrieval performance is then calculated using the proportion of correctly answered questions.

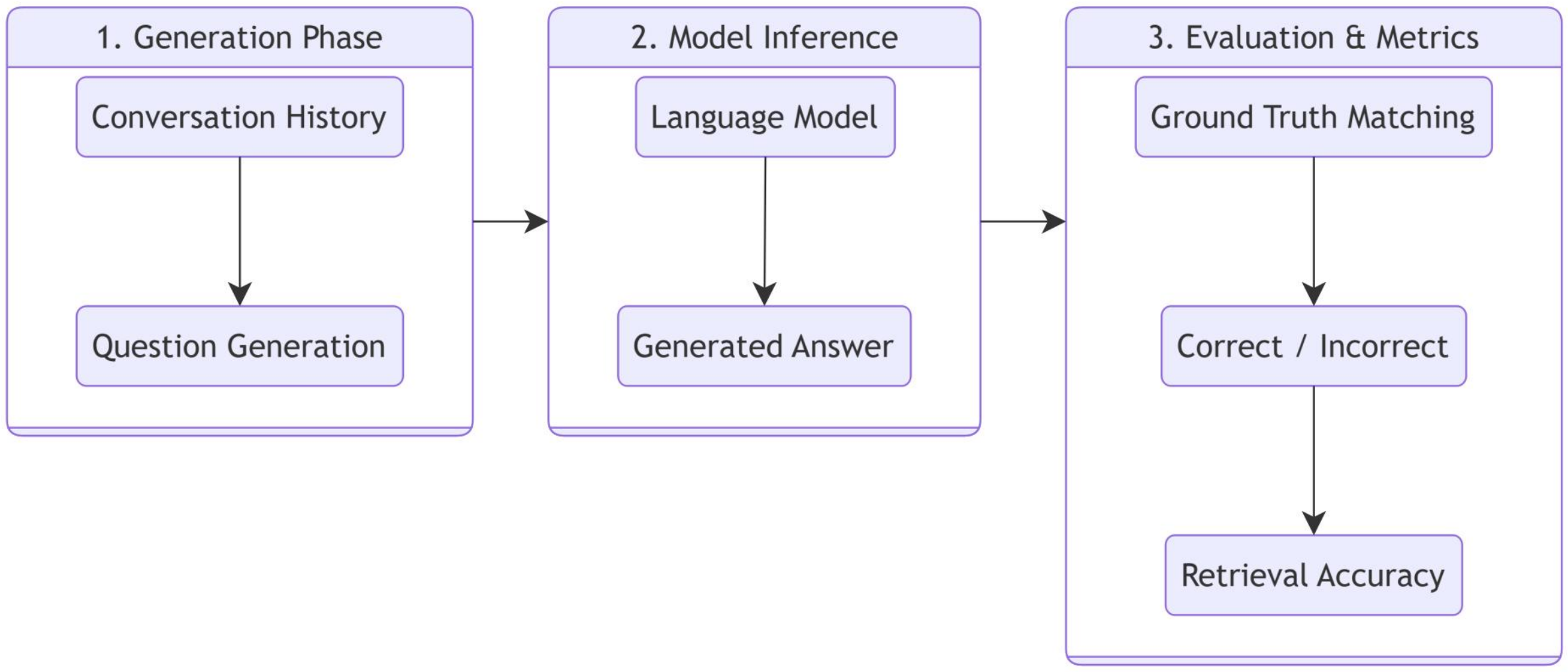

A prediction is considered correct only when the generated answer exactly matches the corresponding ground-truth value. Any missing, incorrect, incomplete, or semantically different response is classified as incorrect. Consequently, retrieval performance is measured using exact-match evaluation without partial credit, ensuring an objective and reproducible scoring process across all evaluated language models.

The final retrieval performance is quantified using Retrieval Accuracy, defined as the percentage of correctly answered retrieval questions over the total number of evaluated questions.

$$\text{Retrieval Accuracy} = \frac{N_{\text{correct}}}{N_{\text{total}}} \times 100\%$$

where $N_{\text{correct}}$ denotes the number of correctly retrieved facts, and $N_{\text{total}}$ represents the total number of retrieval questions evaluated for a given benchmark configuration.

### 3.4.2 Summary Reconstruction Test

The Summary Reconstruction Test (SRT) evaluates a language model's ability to reconstruct previously presented information after the entire conversation context has been processed. Instead of retrieving individual facts through isolated questions, the benchmark prompts the model to extract and reproduce all previously presented factual statements in a structured list. This evaluation measures whether the model can preserve the overall information content rather than recalling individual facts independently.

**Figure 5:** Figure 5 illustrates the evaluation workflow of the Summary Reconstruction Test. After the complete conversation history has been constructed, the language model is instructed to reconstruct every previously presented factual statement as a structured summary. The generated summary is subsequently compared against the benchmark ground-truth facts through automatic line-by-line matching, producing the Critical Information Recall (CIR) score that represents the proportion of correctly reconstructed information.

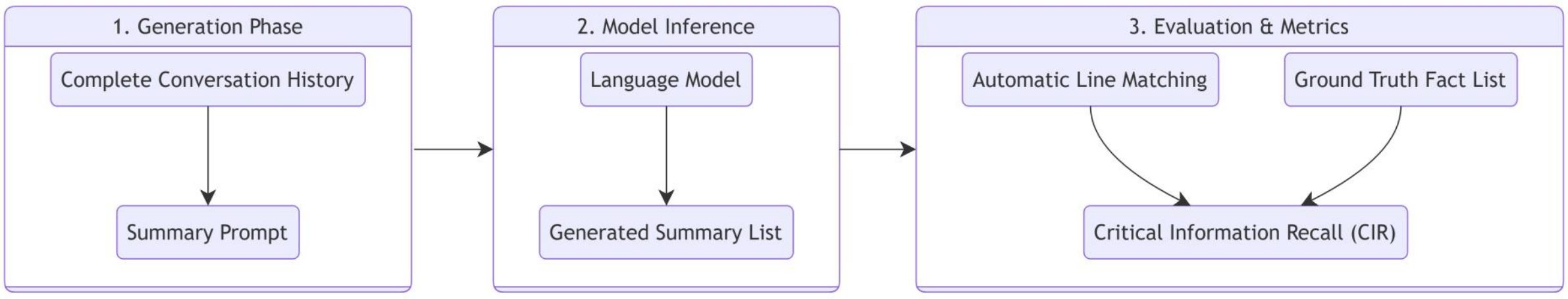

Unlike the Sequential Retrieval Test, which evaluates each fact independently through dedicated retrieval questions, the Summary Reconstruction Test requires the language model to reconstruct the complete set of stored information within a single response. Consequently, this evaluation reflects the model's ability to preserve and organize multiple facts simultaneously under increasing context sizes, providing a complementary perspective on short-term memory behavior.

Each reconstructed statement is compared against its corresponding ground-truth fact using exact-match evaluation. A fact is considered correctly reconstructed only when the generated statement preserves the same subject, attribute, and value as the reference annotation. Statements containing missing, incorrect, incomplete, or semantically different information are classified as incorrect. This deterministic evaluation enables objective measurement without assigning partial credit.

The reconstruction performance is quantified using the Critical Information Recall (CIR) score, defined as the percentage of correctly reconstructed facts relative to the total number of benchmark information items.

$$\text{CIR} = \frac{N_{\text{reconstructed}}}{N_{\text{total}}} \times 100\%$$

where $N_{\text{reconstructed}}$ denotes the number of correctly reconstructed facts, and $N_{\text{total}}$ represents the total number of information items presented during the benchmark.

### 3.4.3 Computational Metrics

In addition to evaluating memory performance, the proposed SRRM benchmark records several computational metrics throughout every benchmark execution. These measurements complement the retrieval and reconstruction evaluations by characterizing the computational cost associated with processing increasing context sizes. Since all models are evaluated under identical hardware and inference configurations, the recorded computational metrics provide a consistent basis for comparing efficiency across different model sizes.

To capture the computational characteristics of each evaluation, the SRRM benchmark measures five complementary metrics throughout the inference process. These metrics quantify the amount of processed information, response generation behavior, execution latency, and hardware resource utilization. Together, they provide a comprehensive view of the computational requirements associated with each evaluated model, complementing the memory-oriented evaluation metrics introduced in the previous subsections.

**Table 12:** The SRRM benchmark records multiple computational metrics alongside memory performance to provide a more comprehensive evaluation of language models. These metrics quantify computational efficiency by measuring context size, response length, inference latency, and hardware resource utilization throughout the benchmark. Together, they complement the retrieval- and reconstruction-based evaluations by revealing the computational trade-offs associated with increasing model parameters and context sizes.

| **Metric** | **Unit** | **Description** |
|---|---|---|
| Input Tokens | Tokens | Number of tokens contained in the input context |
| Output Tokens | Tokens | Number of tokens generated by the language model |
| Inference Time | Seconds | Total response generation time for each evaluation |
| Peak RAM Usage | MB | Maximum system memory consumed during inference |
| Peak VRAM Usage | MB | Maximum GPU memory consumed during inference |

The computational metrics summarized in **Table 12** are collected automatically during every benchmark execution without affecting the evaluation procedure. Unlike Retrieval Accuracy and Critical Information Recall, which measure memory performance, these metrics quantify the computational resources required to perform the benchmark. Consequently, the SRRM benchmark provides both effectiveness and efficiency perspectives, enabling language models to be compared not only by their memory capability but also by their computational requirements under identical experimental conditions.

Collectively, the evaluation methodology presented in this section provides a comprehensive assessment of short-term memory behavior in language models. The Sequential Retrieval Test measures the accurate recall of individual facts, whereas the Summary Reconstruction Test evaluates the preservation of complete information through structured reconstruction. These memory-oriented evaluations are further complemented by computational metrics that characterize inference efficiency and hardware resource utilization. Together, they establish a standardized and reproducible evaluation framework that forms the methodological foundation of the proposed SRRM benchmark.

# 4 Experimental Setup

This section describes the experimental procedure used to evaluate the proposed SRRM benchmark across the Qwen2.5 model family. All experiments were conducted under a standardized evaluation protocol to ensure fair and reproducible comparisons between models. Each language model was evaluated using the same synthetic dataset, identical context sizes, consistent inference configurations, and the same retrieval and reconstruction procedures described in the previous section.

The experimental setup was designed to isolate the effect of model size on short-term memory performance by minimizing variations unrelated to the evaluated models. Consequently, any observed differences in retrieval accuracy, reconstruction capability, and computational efficiency can be primarily attributed to differences in model capacity rather than inconsistencies in the evaluation process.

## 4.1 Experimental Protocol

All experiments were conducted following a standardized protocol to ensure fair, consistent, and reproducible comparisons across all evaluated language models. Every experiment employed the identical benchmark dataset, inference configuration, evaluation procedure, and scoring criteria. Consequently, the only independent variable throughout the study was the model size, allowing performance differences to be attributed primarily to variations in model capacity.

The benchmark protocol consists of the following stages:

1. **Context scaling:** Each evaluated model was tested using five predefined context sizes containing 10, 25, 50, 75, and 100 information items. The information order remained identical for every experiment, ensuring that all models received exactly the same benchmark inputs.

2. **Dataset consistency:** The synthetic benchmark dataset was generated only once and reused throughout the entire evaluation process. No random sampling or dataset regeneration was performed after benchmark construction, guaranteeing identical evaluation data for all models.

3. **Conversation construction:** Information items were sequentially inserted into the conversation history according to the deterministic Round-Robin organization described in Section 3.3. After the complete context had been constructed, the conversation remained unchanged throughout the subsequent evaluation stages.

4. **Sequential Retrieval Test:** One retrieval question was generated for each benchmark record and evaluated in reverse chronological order. Every response was automatically compared against its corresponding ground-truth annotation using exact-match evaluation.

5. **Summary Reconstruction Test:** Following the completion of the retrieval evaluation, the language model was instructed to reconstruct all previously presented information within a single structured response. The reconstructed facts were subsequently evaluated against the benchmark ground truth using the Critical Information Recall (CIR) metric.

6. **Computational measurement:** During every benchmark execution, the system automatically recorded input tokens, output tokens, inference time, peak RAM usage, and peak VRAM usage without modifying the evaluation workflow.

All experiments were executed using the same benchmark implementation and inference configuration throughout the study. By maintaining identical evaluation conditions for every model and context size, the proposed protocol minimizes uncontrolled experimental variation and ensures that the reported results reflect differences in model behavior rather than inconsistencies in the evaluation procedure.

## 4.2 Evaluation Pipeline

The proposed SRRM benchmark employs a fully automated evaluation pipeline designed to ensure consistent execution across all experiments. Every evaluated language model follows the same sequence of operations, from benchmark initialization to final result generation, eliminating procedural variations that could influence the evaluation outcome. This standardized workflow enables fair comparisons between models while maintaining reproducibility throughout the entire benchmarking process.

The evaluation pipeline consists of several sequential stages. For each evaluated model, the benchmark iterates through every predefined context size and constructs the corresponding conversation history from the synthetic dataset. Once the context has been established, the benchmark executes the Sequential Retrieval Test followed by the Summary Reconstruction Test using the identical conversation history. Throughout both evaluation stages, computational metrics are continuously recorded, and all generated outputs are automatically compared against the benchmark ground truth. Upon completion of each evaluation level, the benchmark aggregates the collected metrics into a structured report before proceeding to the next context size and repeating the same procedure for the remaining evaluated models.

**Figure 6:** Figure 6 illustrates the complete execution workflow implemented in the SRRM benchmark. Beginning with model selection, the benchmark evaluates every predefined context size by constructing the conversation history, executing both memory-oriented evaluation tasks, automatically computing performance metrics, and recording computational statistics. The pipeline is repeated until all context sizes have been evaluated for the current model before continuing with the remaining language models. This deterministic execution workflow ensures identical evaluation conditions throughout the benchmark, thereby supporting fair and reproducible model comparisons.

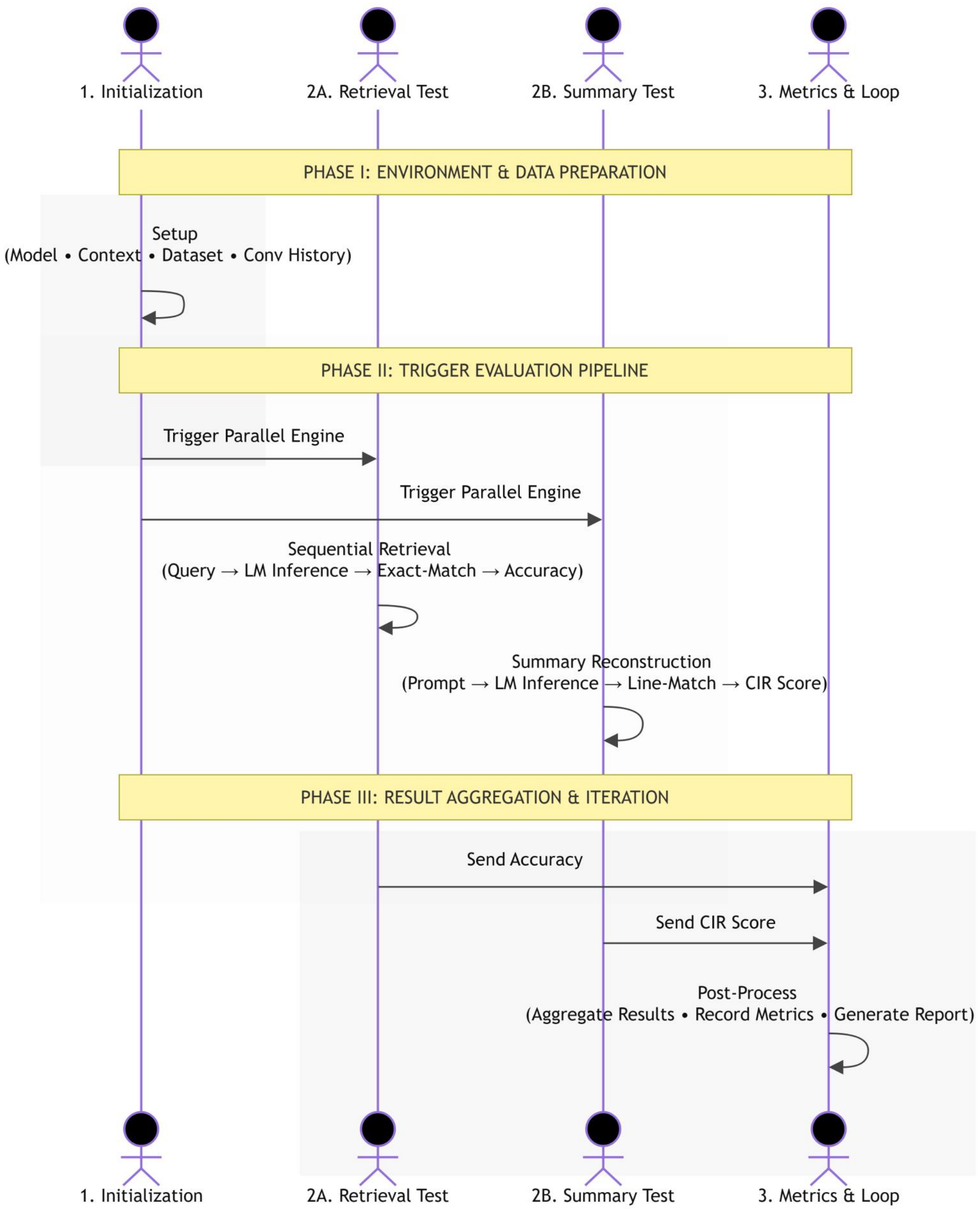

The execution pipeline was designed to minimize manual intervention while preserving deterministic evaluation behavior across all benchmark configurations. By applying an identical workflow to every evaluated model and context size, the benchmark guarantees that the reported retrieval performance, reconstruction capability, and computational efficiency are measured under consistent experimental conditions. Consequently, differences observed in the experimental results can be attributed primarily to model behavior rather than variations in the evaluation procedure.

# 5 Result

This chapter presents the experimental results obtained using the proposed SRRM benchmark across the Qwen2.5 model family. All experiments were conducted under the standardized protocol described in Chapter 4, ensuring identical datasets, inference settings, and evaluation procedures for every model. The reported results include both memory performance and computational measurements across five progressively increasing context sizes.

The results are organized into four sections. The first two sections present the Sequential Retrieval Test and Summary Reconstruction Test independently, followed by a comparative analysis of both evaluation tasks. The final section summarizes the computational performance recorded during benchmark execution. Interpretation of the observed behaviors and discussion of their implications are intentionally reserved for Chapter 6 to maintain a clear separation between experimental observations and analytical interpretation.

## 5.1 Sequential Retrieval Performance

The Sequential Retrieval Test evaluates the ability of each language model to retrieve previously presented information after receiving progressively larger conversational contexts. Retrieval accuracy was measured using exact-match evaluation against the corresponding ground-truth records, with context sizes increasing from 10 to 100 information items. Figure 7 illustrates the retrieval accuracy obtained by each evaluated model across all context levels.

**Figure 7:** As shown in Figure 7, retrieval performance varies across the evaluated models as the context size increases. The figure presents the overall accuracy trend for Qwen2.5-0.5B, Qwen2.5-1.5B, and Qwen2.5-3B under identical benchmark conditions. The numerical retrieval accuracy corresponding to each evaluation level is summarized in Table 13.

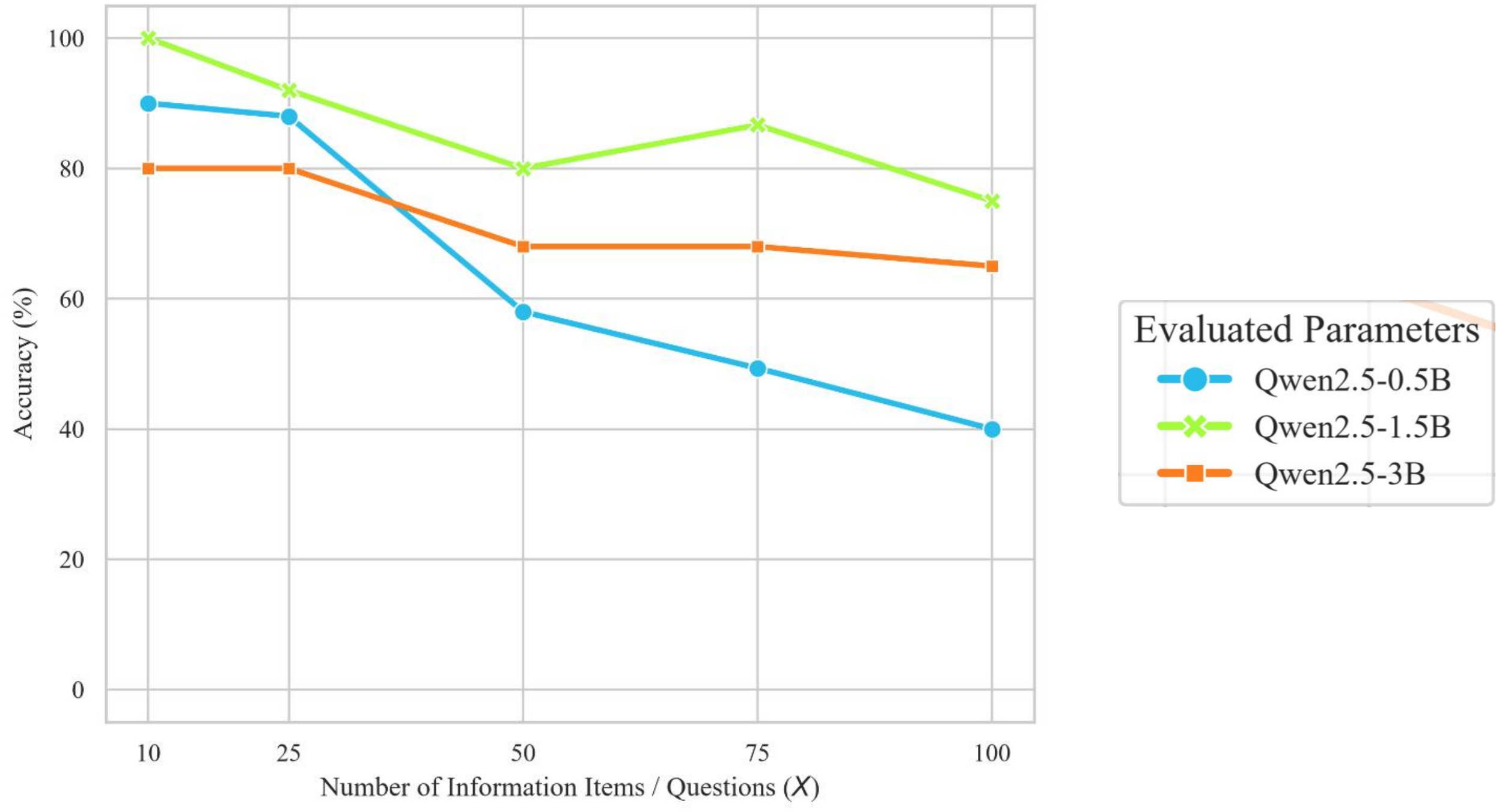

**Table 13:** reports the exact retrieval accuracy obtained by each model at every evaluated context size. Presenting the numerical results in tabular form complements the graphical representation in Figure 7 by providing precise values for each experimental condition, enabling direct comparison across both models and context levels.

| Context Size | Qwen2.5-0.5B | Qwen2.5-1.5B | Qwen2.5-3B |
|---|---|---|---|
| 10 | 90,0% | 100,0% | 80,0% |
| 25 | 88,0% | 92,0% | 80,0% |
| 50 | 58,0% | 80,0% | 68,0% |
| 75 | 49,3% | 86,7% | 68,0% |
| 100 | 40,0% | 75,0% | 65,0% |

To provide an overall summary of retrieval behavior, Table 14 reports several descriptive statistics computed from the results presented in Table 13. These statistics summarize the average retrieval performance of each model, together with the highest and lowest observed accuracies and the overall decrease in retrieval accuracy between the smallest and largest evaluated context sizes.

**Table 14:** summarizes the overall retrieval characteristics of each evaluated model. Mean Accuracy represents the average retrieval accuracy across all five context sizes, Max Accuracy and Min Accuracy denote the highest and lowest recorded retrieval performance, respectively, while Accuracy Drop measures the absolute decrease in retrieval accuracy between the evaluations conducted with 10 and 100 information items.

| Model | Mean Acc | Max Acc (@10) | Min Acc (@100) | Acc Drop |
|---|---|---|---|---|
| **Qwen2.5-0.5B** | 65,1% | 90,0% | 40,0% | 50,0% |
| **Qwen2.5-1.5B** | 86,7% | 100,0% | 75,0% | 25,0% |
| **Qwen2.5-3B** | 72,2% | 80,0% | 65,0% | 15,0% |

Overall, the Sequential Retrieval Test provides a quantitative comparison of retrieval performance across progressively increasing context sizes. The complete retrieval results presented in this section serve as the basis for the Summary Reconstruction evaluation presented in the following section, where the models are assessed using a complementary memory task based on information reconstruction rather than direct retrieval.

## 5.2 Summary Reconstruction Performance

The Summary Reconstruction Test evaluates the ability of each language model to reconstruct previously presented information after the complete conversational context has been processed. Unlike the Sequential Retrieval Test, which measures direct retrieval of individual facts, this evaluation assesses the preservation of critical information through a single generated summary. Reconstruction performance was quantified using the proposed Critical Information Recall (CIR) metric across five progressively increasing context sizes.

During this evaluation, each model was instructed to generate a comprehensive summary after all benchmark information had been presented within the conversation. The generated summary was then compared against the corresponding structured ground-truth annotations using the Critical Information Recall metric described in Section 3.4.2. This evaluation measures the extent to which essential information is preserved during summary generation rather than the ability to retrieve isolated facts through explicit questioning.

Figure 8 presents the reconstruction performance of the evaluated Qwen2.5 models across all benchmark levels. The figure illustrates how the measured CIR scores change as the amount of contextual information increases, providing an overall view of reconstruction performance under progressively larger conversational contexts.

**Figure 8 :** As illustrated in Figure 8, reconstruction performance changes across the evaluated context sizes for all three language models. The figure summarizes the overall reconstruction trend measured using the proposed CIR metric under identical experimental conditions. The corresponding numerical CIR scores for each context level are presented in Table 15.

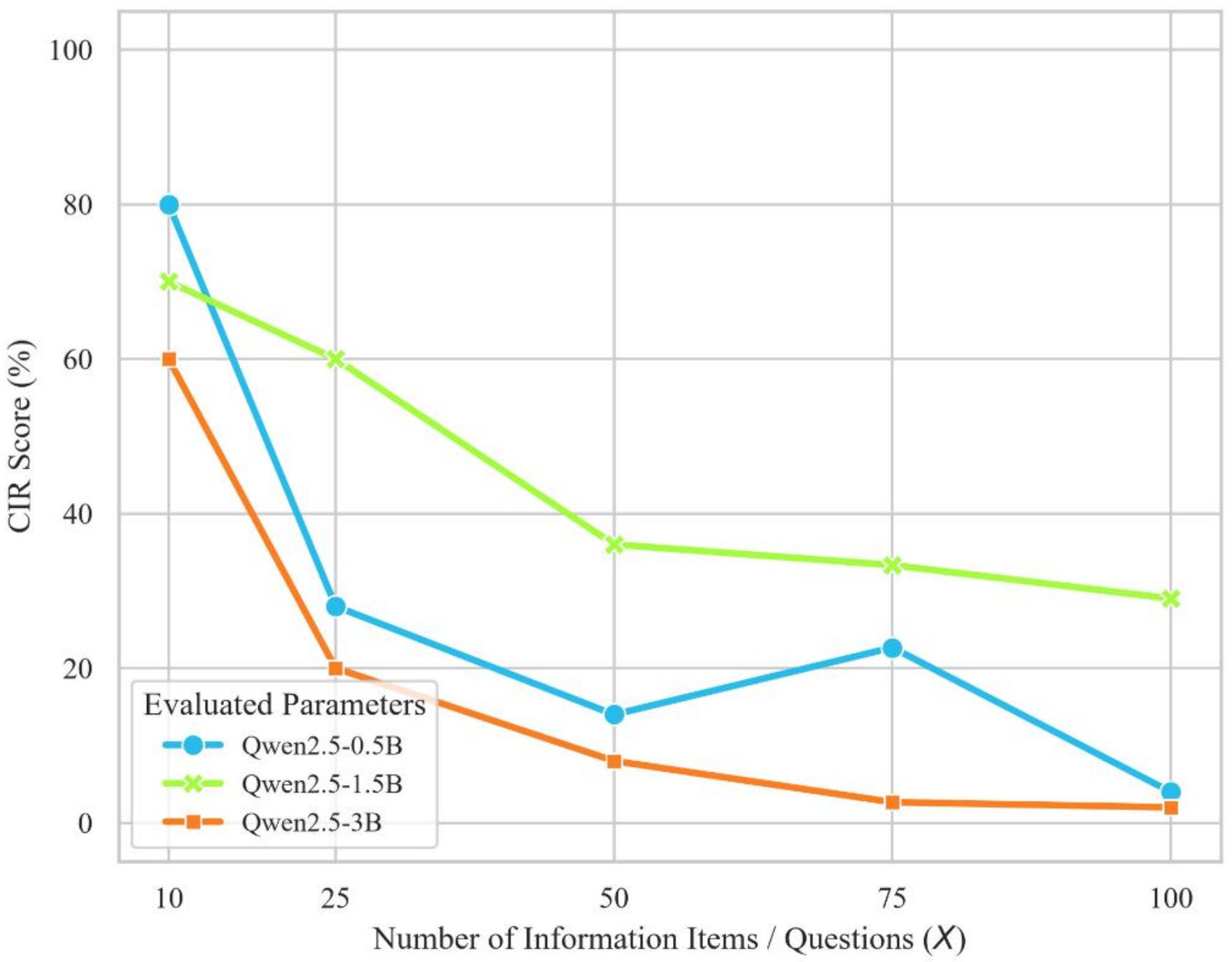

**Table 15 :** Reports the exact CIR scores obtained by each evaluated model at every benchmark level. The tabulated values complement the graphical representation shown in Figure 8 by providing precise reconstruction scores for each context size, facilitating direct comparison across different parameter scales.

| Context Size | Qwen2.5-0.5B | Qwen2.5-1.5B | Qwen2.5-3B |
|---|---|---|---|
| 10 | 80,0% | 70,0% | 60,0% |
| 25 | 28,0% | 60,0% | 20,0% |
| 50 | 14,0% | 36,0% | 8,0% |
| 75 | 22,7% | 33,3% | 2,7% |
| 100 | 4,0% | 29,0% | 2,0% |

To provide an overall summary of reconstruction performance, Table 16 presents descriptive statistics derived from the CIR scores reported in Table 15. These statistics summarize the average reconstruction capability of each model together with the highest and lowest recorded CIR scores and the overall decrease observed between the smallest and largest evaluated context sizes.

**Table 16 :** Summarizes the overall reconstruction characteristics of each evaluated model. Mean CIR represents the average reconstruction score across all five context sizes, Max CIR and Min CIR indicate the highest and lowest reconstruction scores observed during the benchmark, respectively, while CIR Drop denotes the absolute decrease in reconstruction performance between the evaluations conducted using 10 and 100 information items.

| Model | Mean CIR | Max CIR (@10) | Min CIR (@100) | CIR Drop |
|---|---|---|---|---|
| **Qwen2.5-0.5B** | 29,7% | 80,0% | 4,0% | 76,0% |
| **Qwen2.5-1.5B** | 45,7% | 70,0% | 29,0% | 41,0% |
| **Qwen2.5-3B** | 18,5% | 60,0% | 2,0% | 58,0% |

Overall, the Summary Reconstruction Test provides a quantitative evaluation of the models' ability to preserve critical information as conversational context increases. Together with the retrieval results presented in the previous section, these findings establish the complete set of experimental observations used for the comparative analysis presented in the following section, where both memory evaluation tasks are examined side by side.

## 5.3 Comparative Analysis

Following the independent evaluations presented in Sections 5.1 and 5.2, this section provides a comparative analysis of retrieval and reconstruction performance across the evaluated Qwen2.5 models. Rather than considering both evaluation tasks separately, the results are examined collectively to provide a comprehensive overview of memory behavior under the proposed SRRM benchmark. The comparison includes overall benchmark performance, category-level retrieval characteristics, reconstruction behavior, and an aggregated performance ranking across all evaluated models.

To facilitate direct comparison between both memory evaluation tasks, Figure 9 summarizes the average retrieval accuracy and average Critical Information Recall (CIR) obtained by each evaluated model. The comparison provides a concise overview of the relationship between explicit information retrieval and summary-based information reconstruction using identical conversational contexts.

**Figure 9:** Overall comparison between mean retrieval accuracy and mean CIR score.

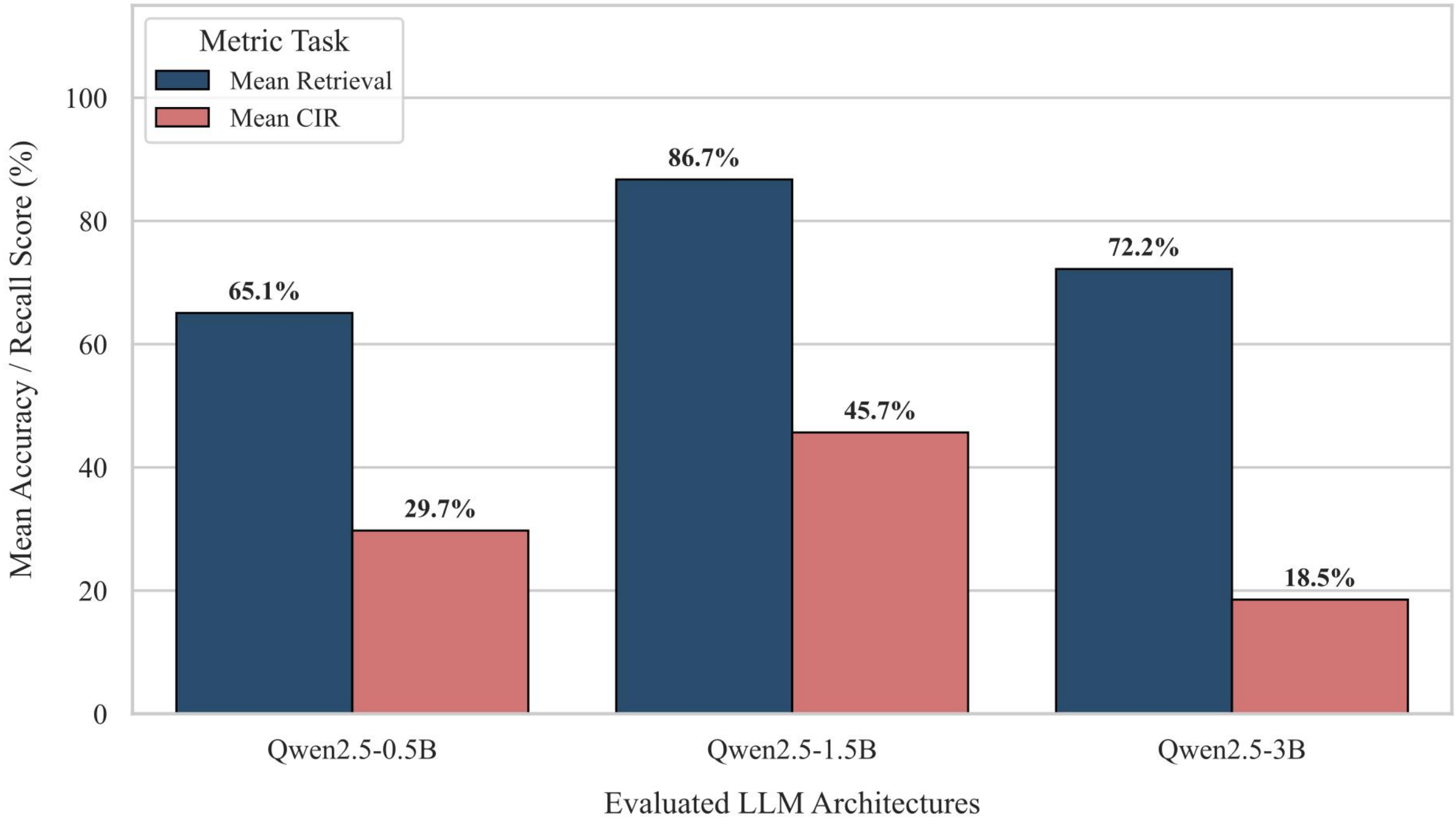

Although the previous comparison summarizes the overall benchmark performance, the retrieval results can also be examined at the category level. Figure 10 presents the average retrieval accuracy obtained for each dataset information category, enabling a more detailed comparison of model performance across different semantic information types included in the SRRM benchmark.

**Figure 10:** Category-specific retrieval accuracy heatmap matrix

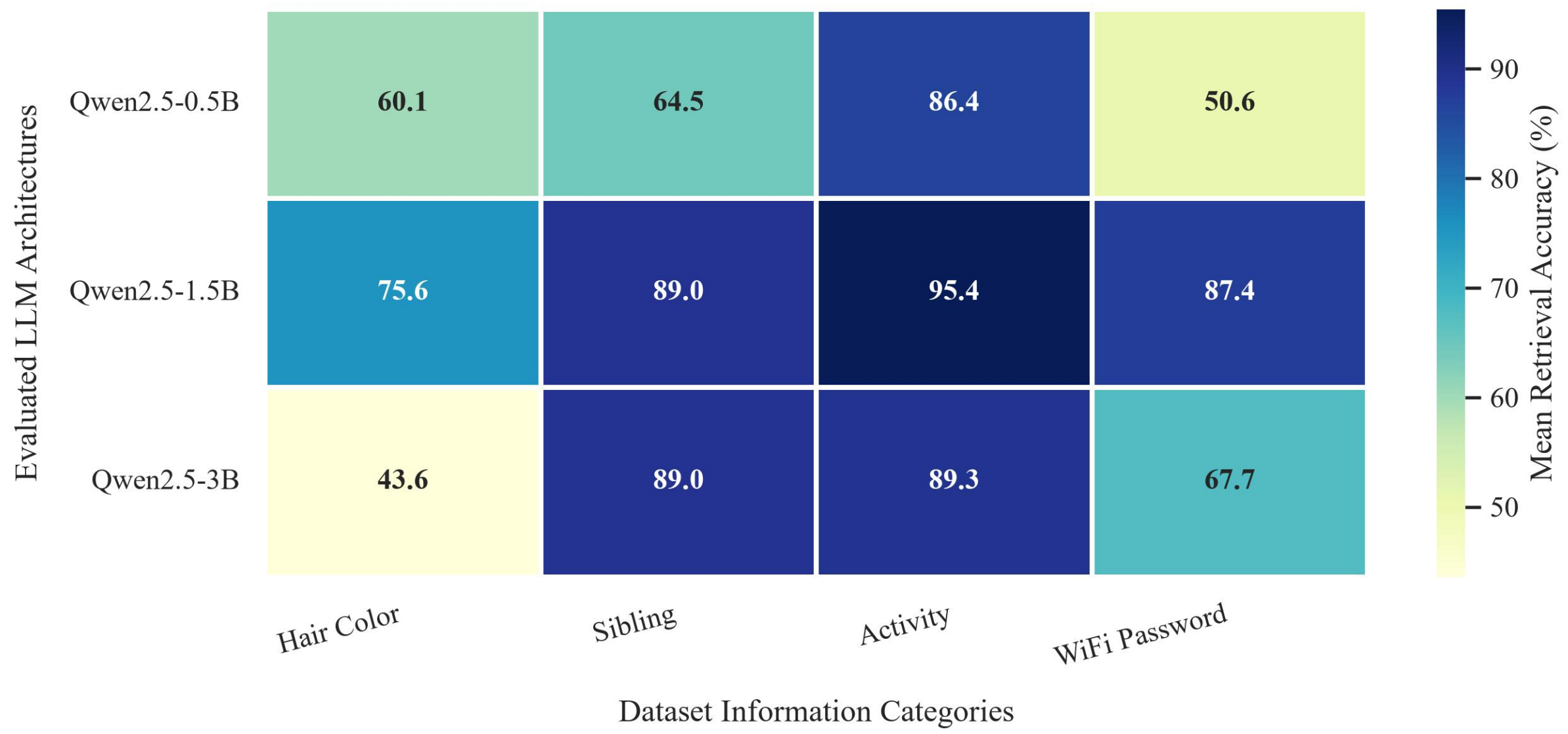

**Figure 10** visualizes the mean retrieval accuracy for each evaluated language model across the four benchmark information categories. Color intensity represents retrieval performance averaged over all evaluated context sizes, while the numerical values indicate the corresponding percentage accuracy for each model–category combination.

If category-level reconstruction analysis is available, it can likewise be summarized using the same visualization format to complement the retrieval analysis. Figure 11 presents the corresponding category-wise reconstruction performance measured using the Critical Information Recall metric, allowing reconstruction behavior to be examined under the same information categories used throughout the benchmark.

**Figure 11:** Category-specific reconstruction performance heatmap matrix

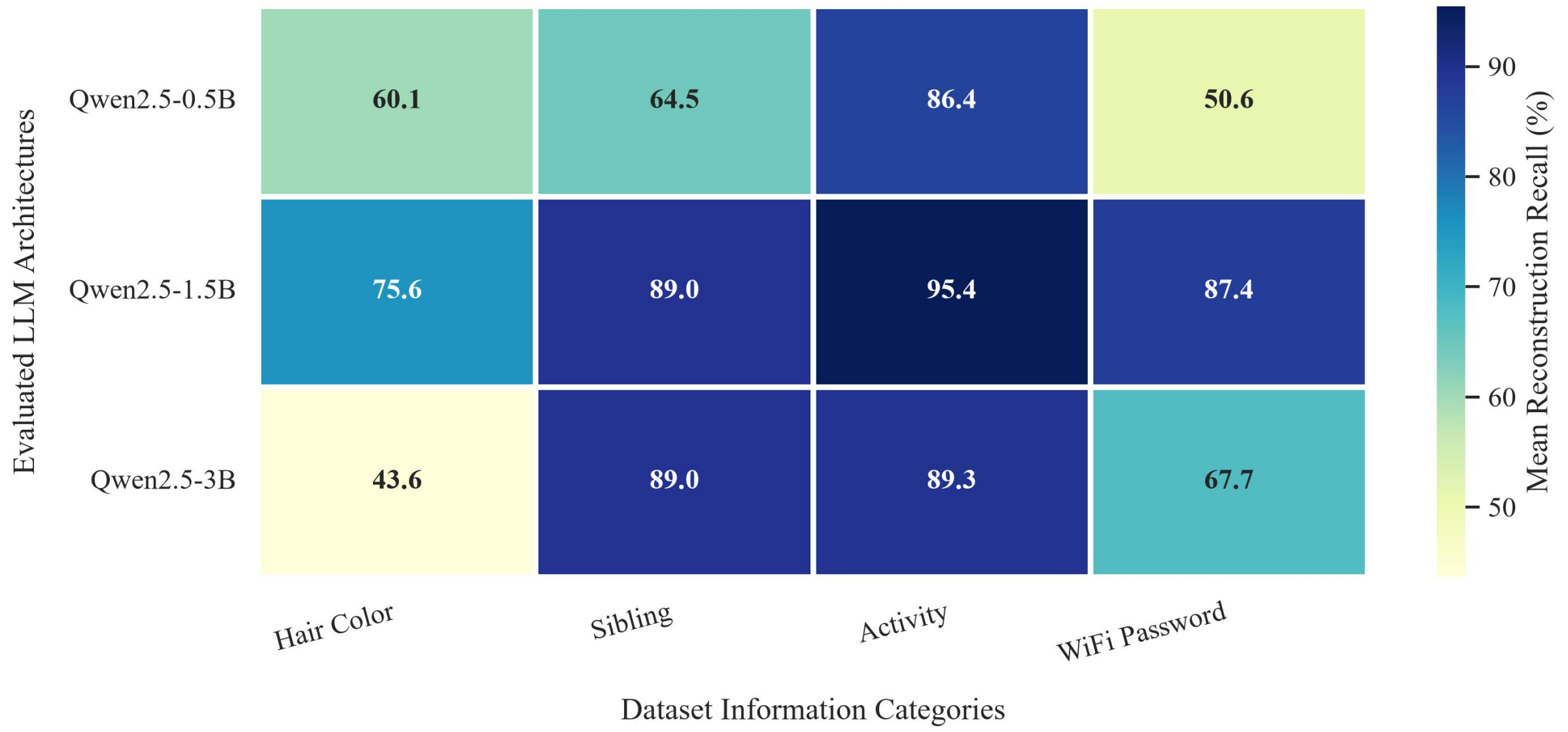

The category-wise analyses are complemented by an aggregated comparison presented in Table 18, which summarizes the overall benchmark ranking for each evaluated model. Rather than focusing on individual evaluation tasks, the ranking combines the principal retrieval and reconstruction metrics into a single benchmark-level comparison to provide an overall performance overview.

**Table 18:**. Overall SRRM benchmark performance ranking

| Model | Mean Ret. | Mean CIR | Ret./CIR | Overall | Rank |
|---|---|---|---|---|---|
| **Qwen2.5-0.5B** | 65,1% | 29,7% | 2,19 | **47,4%** | **2** |
| **Qwen2.5-1.5B** | 86,7% | 45,7% | 1,90 | **66,2%** | **1** |
| **Qwen2.5-3B** | 72,2% | 18,5% | 3,90 | **45,4%** | **3** |

**Table 18** summarizes the benchmark-level ranking derived from the retrieval and reconstruction evaluations presented in the preceding sections. The ranking provides a consolidated comparison of overall memory performance across the evaluated models while preserving the detailed retrieval, reconstruction, and category-level analyses reported previously. As such, it serves as a concise reference for comparing the relative benchmark performance of the evaluated Qwen2.5 model family.

Taken together, the results presented in this section integrate the principal findings obtained from both memory evaluation tasks into a unified benchmark perspective. The overall comparison, category-level analyses, and benchmark ranking collectively provide a comprehensive summary of the experimental observations reported in this chapter. These findings establish the empirical foundation for the subsequent discussion, where the observed performance characteristics, potential explanations, and broader implications of the proposed SRRM benchmark are examined in greater detail.

# 6 Discussion

The experimental results presented in Chapter 5 demonstrate distinct memory behaviors across the evaluated Qwen2.5 models under the proposed SRRM benchmark. While the retrieval and reconstruction evaluations provide quantitative measurements of memory performance, the reported results alone do not explain the underlying patterns observed throughout the experiments. This chapter therefore interprets the principal findings by examining the relationships between retrieval performance, summary reconstruction capability, model scaling, and computational efficiency.

Rather than introducing additional benchmark evaluations, the discussion synthesizes the results obtained from the Sequential Retrieval Test, the Summary Reconstruction Test, and the computational measurements to identify broader behavioral patterns across the evaluated models. Where appropriate, supplementary analyses are presented to further investigate unexpected observations that emerged during the benchmark, providing additional evidence without altering the standardized SRRM evaluation protocol.

The discussion is organized into five sections. The first section examines the relationship between retrieval and reconstruction performance, followed by an analysis of model scaling behavior and category-level memory characteristics. The subsequent sections discuss the implications of the proposed benchmark for memory evaluation in large language models, together with the limitations of the current study and directions for future research.

## 6.1 Analysis of Retrieval-Reconstruction Behavior

The experimental results presented in Chapter 5 reveal a consistent discrepancy between retrieval performance and summary reconstruction performance across all evaluated language models. Although both evaluation tasks measure memory-related capabilities using the same conversational contexts, they assess fundamentally different aspects of information processing. The Sequential Retrieval Test evaluates the ability to explicitly recall previously presented facts through direct questioning, whereas the Summary Reconstruction Test measures the preservation of critical information when the model is required to generate a coherent summary from the accumulated conversational context. Consequently, comparing the outcomes of these two evaluations provides insight into how effectively retrieved information is retained during higher-level language generation.

To facilitate this comparison, **Figure 13** summarizes the mean retrieval accuracy and the corresponding mean Critical Information Recall (CIR) score obtained by each evaluated model. By presenting both measurements within a single visualization, the figure highlights the relationship between explicit information retrieval and summary-based information reconstruction while illustrating the magnitude of the performance gap observed throughout the benchmark.

**Figure 13:** Demonstrates that all evaluated models achieved substantially higher retrieval performance than reconstruction performance. Although the magnitude of this difference varies across the Qwen2.5 model family, the same overall pattern is consistently observed, indicating that successful retrieval of previously presented information does not necessarily translate into equivalent preservation of that information during summary generation. This observation suggests that retrieval and reconstruction represent related but distinct memory capabilities, supporting the motivation for evaluating both tasks independently within the proposed SRRM benchmark.

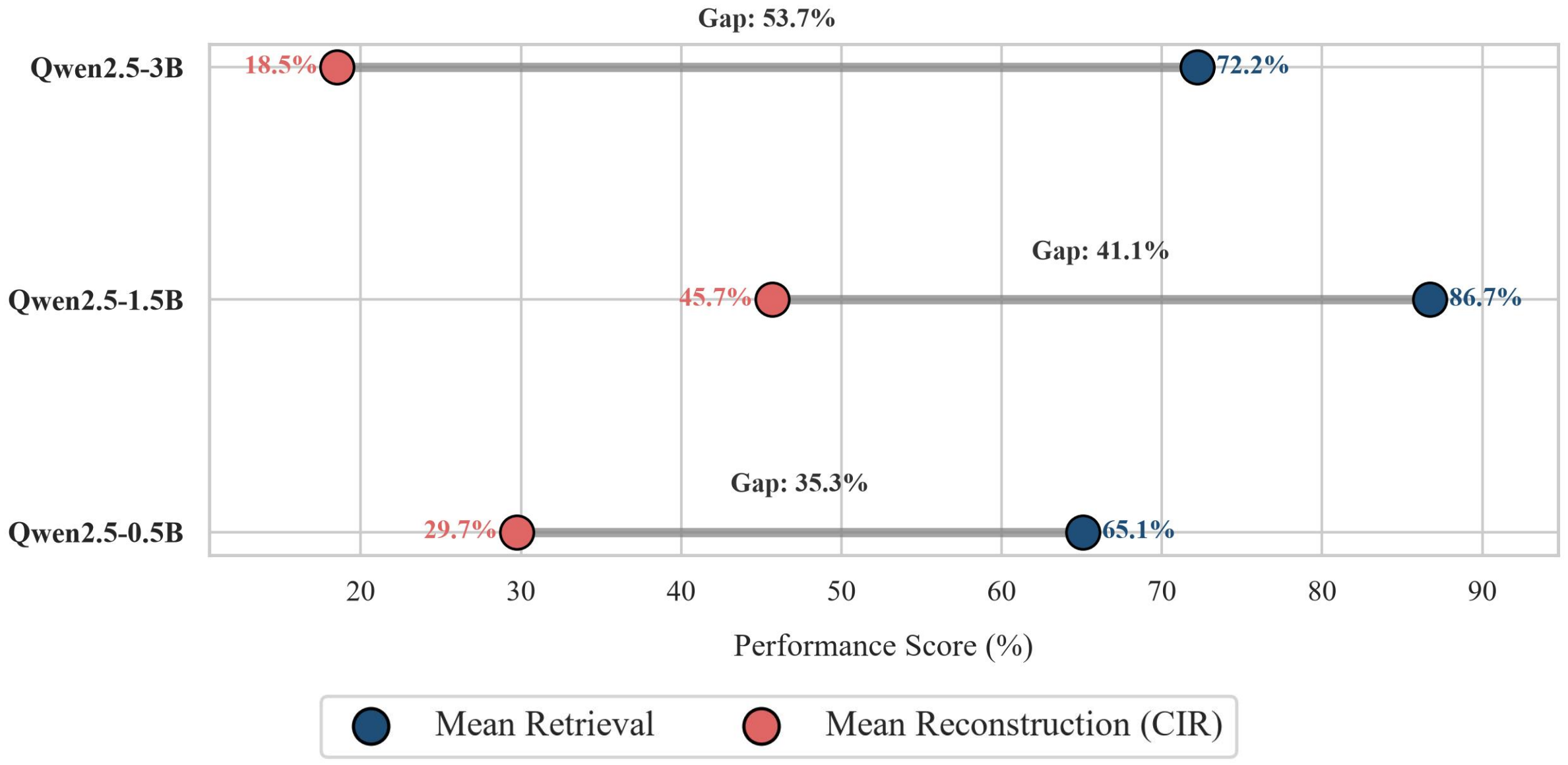

The performance gap illustrated in **Figure 13** is quantified in Table 19 using both absolute and relative gap measurements. While the absolute gap measures the numerical difference between mean retrieval accuracy and mean CIR, the relative gap normalizes this difference with respect to retrieval performance, allowing comparisons across models with different overall accuracy levels.

**Table 19:** Retrieval–reconstruction gap analysis

| Model | Mean Ret. | Mean CIR | Abs. Gap | Rel. Gap (%) |
|---|---|---|---|---|
| **Qwen2.5-0.5B** | 65,1% | 29,7% | 35,3% | **54,3%** |
| **Qwen2.5-1.5B** | 86,7% | 45,7% | 41,1% | **47,3%** |
| **Qwen2.5-3B** | 72,2% | 18,5% | 53,7% | **74,3%** |

**Table 19** indicates that the observed differences extend beyond simple variations in retrieval accuracy. Although Qwen2.5-3B maintains a relatively competitive retrieval performance, it exhibits the largest absolute and relative gap among the evaluated models, whereas Qwen2.5-1.5B records the smallest relative gap despite achieving the highest retrieval accuracy. These findings indicate that higher retrieval performance alone does not necessarily correspond to stronger information preservation during summary reconstruction.

To further characterize this relationship, an additional descriptive indicator, referred to as the **Memory Preservation Ratio (MPR)**, was calculated. MPR is defined as the ratio between the mean CIR score and the mean retrieval accuracy, representing the proportion of successfully retrieved information that is preserved during summary generation. Unlike the evaluation metrics introduced in the SRRM benchmark, MPR is employed solely as an analytical measure within the discussion to facilitate interpretation of the observed benchmark results.

**Table 20:** Memory Preservation Ratio (MPR) analysis

| Model | Mean Ret. | Mean CIR | MPR | Interpretation |
|---|---|---|---|---|
| **Qwen2.5-0.5B** | 65,1% | 29,7% | **0,46** | **Moderate** |
| **Qwen2.5-1.5B** | 86,7% | 45,7% | **0,53** | **High** |
| **Qwen2.5-3B** | 72,2% | 18,5% | **0,26** | **Low** |

**Table 20** shows that the evaluated models exhibit noticeably different levels of memory preservation despite being assessed under identical benchmark conditions. Among the three models, Qwen2.5-1.5B achieves the highest MPR, indicating that more than half of the information successfully retrieved during direct questioning is also retained during summary reconstruction. In contrast, Qwen2.5-3B records the lowest MPR, suggesting a substantially larger reduction in preserved information when transitioning from explicit retrieval to summary generation. Qwen2.5-0.5B occupies an intermediate position between these two models.

Taken together, the analyses presented in this section demonstrate that retrieval accuracy alone is insufficient to fully characterize memory performance in large language models. Models exhibiting comparable retrieval capability may differ substantially in their ability to preserve retrieved information during summary generation, indicating that retrieval and reconstruction capture complementary aspects of long-context memory behavior. This distinction provides additional evidence supporting the dual-evaluation design of the proposed SRRM benchmark, while also motivating a more detailed examination of model scaling behavior in the following section.

## 6.2 Analysis of Model Scaling

Scaling laws have consistently demonstrated that increasing model size generally leads to improved performance across a wide range of language understanding and reasoning tasks. Under this expectation, larger language models are typically assumed to possess greater representational capacity, enabling them to process more complex contextual information and achieve higher task performance. Consequently, one would reasonably expect the Qwen2.5-3B model to outperform both the 1.5B and 0.5B variants across all memory evaluations included in the proposed SRRM benchmark.

However, the experimental findings presented in Chapter 5 reveal a substantially different scaling behavior. Rather than exhibiting a monotonic relationship between parameter count and benchmark performance, the evaluated models demonstrated a non-linear performance pattern in which the intermediate-sized Qwen2.5-1.5B model consistently achieved the strongest results. This superiority was not limited to a single evaluation task, but instead appeared repeatedly across the Sequential Retrieval Test, the Summary Reconstruction Test, and the Memory Preservation Ratio (MPR). Such consistency suggests that the observed phenomenon cannot be attributed to an isolated benchmark metric or a single experimental scenario.

**Key Observation**

Across all principal benchmark indicators, including retrieval accuracy, reconstruction accuracy, and Memory Preservation Ratio (MPR), the Qwen2.5-1.5B model consistently outperformed the larger Qwen2.5-3B model. The recurrence of this ranking across multiple independent evaluation procedures suggests that the observed behavior reflects an intrinsic characteristic of the evaluated models rather than an isolated measurement artifact.

The consistency of these findings is particularly noteworthy because the three principal benchmark indicators evaluate different aspects of long-context memory. While the Sequential Retrieval Test measures explicit recall of previously presented information, the Summary Reconstruction Test evaluates the ability to preserve and reorganize distributed information into a coherent summary. The Memory Preservation Ratio (MPR) further integrates these two perspectives by quantifying how effectively retrieved information is retained during summary generation. Since all three evaluation procedures independently produced the same performance ranking, the observed phenomenon is unlikely to result from metric-specific characteristics alone.

Furthermore, the experimental protocol was intentionally designed to minimize potential sources of external variation. All evaluated models were tested using the identical benchmark dataset, prompt templates, context organization strategy, retrieval sequence, reconstruction procedure, inference configuration, and computational environment. Consequently, differences in dataset composition, prompt engineering, evaluation methodology, or hardware configuration are unlikely to account for the unexpected superiority of the Qwen2.5-1.5B model.

After considering these controlled conditions, several hypotheses emerge as plausible explanations for the observed scaling behavior.

Hypothesized Explanations

**H1 – Context Utilization Efficiency**

The Qwen2.5-1.5B model may utilize long-context information more efficiently, allowing relevant facts to remain accessible throughout both retrieval and summary reconstruction.

**H2 – Memory Representation Strategy**

The evaluated models may differ in how contextual information is internally represented and maintained, resulting in different levels of information preservation as conversational context increases.

The larger Qwen2.5-3B model may compress accumulated contextual information more aggressively, leading to the loss of fine-grained factual details despite maintaining relatively strong retrieval capability.

**H3 – Information Compression Behavior**

Long-context memory performance may depend on internal architectural or optimization characteristics beyond parameter count alone, causing the observed scaling behavior to deviate from the conventional expectation that larger language models consistently achieve superior memory performance.

**H4 – Scaling-Related Internal Mechanisms**

Although these hypotheses are consistent with the experimental observations, the current SRRM benchmark was designed to measure memory performance rather than identify its underlying causes. Therefore, the available benchmark evidence is insufficient to determine which of the proposed hypotheses most accurately explains the observed scaling behavior. **Investigating these hypotheses requires dedicated experiments specifically designed to isolate the underlying mechanisms of long-context memory, which fall beyond the primary scope of the present study. Consequently, this phenomenon remains an important direction for future research.**

## 6.3 Limitations of the Proposed Benchmark

Although the proposed SRRM benchmark successfully demonstrates its ability to evaluate long-context memory through complementary retrieval and reconstruction tasks, several limitations should be acknowledged. These limitations primarily reflect the controlled design choices adopted during the initial development and validation of the benchmark rather than inherent weaknesses of the evaluation framework itself. Identifying these limitations is important for defining the scope of the present study while providing clear directions for future extensions of the benchmark.

Evaluation Scope. The proposed SRRM benchmark focuses exclusively on evaluating long-context memory preservation. Consequently, it does not assess other language model capabilities such as logical reasoning, mathematical problem solving, planning, tool utilization, factual accuracy, or hallucination detection. This design was intentionally adopted to isolate memory-related behavior from other cognitive abilities, thereby allowing the benchmark to provide a more focused evaluation of information retention. Future benchmark extensions may integrate SRRM with complementary evaluation frameworks to investigate the relationship between long-context memory and broader language model capabilities.

**Limitation 1: Model Coverage**

The experimental validation presented in this study was conducted exclusively using the Qwen2.5 model family, covering parameter scales from 0.5B to 3B. Restricting the evaluation to a single model family reduces architectural variability and enables a more controlled validation of the proposed benchmark during its initial development. Nevertheless, the generalizability of SRRM across different model architectures remains to be investigated. Future studies may extend the benchmark to additional open-source and proprietary language models, including larger parameter scales and architectures developed by different organizations.

**Limitation 2: Dataset Design**

The benchmark utilizes a synthetic dataset specifically designed to provide deterministic ground-truth information for objective evaluation. While this controlled design substantially improves experimental reproducibility and exact-match scoring reliability, it may not fully represent the complexity and variability of naturally occurring human conversations or real-world long-context interactions. Future work may therefore incorporate more realistic conversational datasets, document-based contexts, multilingual scenarios, or agent-based interactions while preserving the benchmark's reproducibility.

**Limitation 3: Evaluation Metrics**

The current benchmark evaluates retrieval performance using exact-match accuracy and reconstruction performance using the Context Information Reconstruction (CIR) metric. Although these metrics provide objective and reproducible measurements of memory preservation, they primarily emphasize factual information retention and reconstruction fidelity. Other evaluation methodologies, such as semantic similarity metrics, embedding-based comparisons, factual consistency evaluation, or LLM-assisted assessment, may provide additional perspectives on long-context memory performance. Integrating multiple complementary evaluation metrics represents a promising direction for future benchmark development.

**Limitation 4: Context Scale**

The present study evaluates memory preservation using benchmark contexts containing between 10 and 100 information items. This range was selected to provide sufficient contextual complexity while maintaining manageable computational requirements during the initial validation of SRRM. However, recent language models increasingly support substantially larger context windows, ranging from hundreds of thousands to millions of tokens. Consequently, future studies may investigate the applicability of SRRM under substantially longer contexts to determine whether the benchmark remains effective for evaluating next-generation long-context language models.

**Limitation 5: Despite these limitations**

They primarily represent opportunities for extending the proposed benchmark rather than deficiencies in its current design. The controlled experimental configuration adopted throughout this study establishes a reliable foundation for evaluating long-context memory while maintaining high reproducibility across different model sizes. As future research expands model coverage, dataset diversity, evaluation metrics, and context scales, SRRM may evolve into a more comprehensive benchmark for systematically evaluating long-context memory preservation across a broader range of large language models and application scenarios.

## 6.4 Implications and Future Work

The findings presented throughout this study demonstrate that long-context memory cannot be sufficiently characterized by retrieval performance alone. While retrieval accuracy indicates whether previously presented information remains accessible, it does not necessarily reflect how effectively that information is preserved when the model is required to reconstruct an entire context. By combining sequential retrieval, summary reconstruction, and the Memory Preservation Ratio (MPR), the proposed SRRM benchmark provides a more comprehensive perspective on long-context memory behavior than either evaluation task alone.

From a practical perspective, SRRM may provide a useful evaluation framework for selecting language models intended for memory-intensive applications. Systems such as conversational assistants, long-term dialogue agents, document understanding systems, retrieval-augmented generation (RAG), and persistent AI agents increasingly rely on the ability to preserve contextual information across extended interactions. Since these applications require both accurate retrieval and consistent information preservation, the benchmark may assist researchers and practitioners in identifying models that maintain long-context memory more reliably under controlled evaluation conditions.

The benchmark also establishes a reproducible experimental protocol that can be readily extended to additional language model families. Although the present study evaluated only the Qwen2.5 model family, the SRRM evaluation pipeline is largely model-independent and may therefore be applied to other open-source or proprietary language models with minimal modification. Future benchmark studies may compare architectural families, investigate larger parameter scales, or evaluate recently developed long-context language models under the same standardized evaluation procedure.

Several opportunities also exist for extending the benchmark itself. Possible future developments include:

- Evaluating substantially longer context windows involving thousands or millions of tokens.
- Incorporating multilingual benchmark datasets to assess memory preservation across different languages.
- Extending the benchmark to multimodal language models capable of processing text together with images, audio, or other modalities.
- Integrating semantic similarity, factual consistency, or embedding-based evaluation methods alongside exact-match scoring.
- Evaluating memory preservation under realistic conversational, document-centric, or agent-based interaction scenarios.

Beyond benchmark development, the unexpected scaling behavior observed in this study also suggests several promising directions for future research. In particular, the consistent superiority of the intermediate-sized Qwen2.5-1.5B model over the larger 3B variant indicates that long-context memory preservation may depend on factors beyond parameter count alone. Although identifying the mechanisms responsible for this phenomenon lies beyond the scope of the present work, the benchmark provides a reproducible foundation for future investigations into context utilization strategies, memory representation, information compression, architectural optimization, and other factors that may influence long-context memory performance.

Overall, the proposed SRRM benchmark is intended not as a fixed evaluation framework, but as an extensible foundation for systematic long-context memory assessment. As language models continue to evolve toward increasingly larger context windows and more memory-intensive applications, reproducible benchmarks that jointly evaluate retrieval accuracy, reconstruction capability, and memory preservation will become increasingly important for understanding and comparing the long-context behavior of future large language models.

# 7 Conclusion

This paper introduced Simple Retrieval–Reconstruction Memory (SRRM), a lightweight and fully reproducible benchmark designed to evaluate long-context memory preservation in Small Language Models (SLMs). Unlike existing retrieval-oriented evaluations, SRRM jointly measures two complementary aspects of memory behavior: Targeted Retrieval, which evaluates the ability to recall individual facts, and Global Information Reconstruction, which evaluates the ability to reconstruct an entire collection of previously presented information. To further integrate these complementary evaluations, this study also proposed the Memory Preservation Ratio (MPR) as a benchmark-level metric that quantifies the relationship between retrieval capability and reconstruction performance. Together, these components establish a systematic framework for evaluating long-context memory from multiple perspectives while maintaining reproducibility across different experimental settings.

The experimental evaluation demonstrated that retrieval accuracy and reconstruction capability exhibit substantially different degradation patterns as contextual information progressively increases. While retrieval performance generally decreased with increasing context load, reconstruction performance deteriorated considerably faster across all evaluated models, indicating that successful retrieval alone is insufficient to characterize long-context memory preservation. Furthermore, the benchmark consistently revealed that the intermediate-sized Qwen2.5-1.5B model achieved superior performance over the larger Qwen2.5-3B model across retrieval accuracy, reconstruction performance, and Memory Preservation Ratio. This unexpected scaling behavior suggests that parameter count alone does not fully explain long-context memory preservation and highlights the importance of complementary evaluation strategies.

**RQ1. How does retrieval performance change as contextual information progressively increases in small language models?**

Retrieval performance consistently decreased as contextual information increased across all evaluated models. However, the magnitude of degradation differed considerably between model sizes. The Qwen2.5-1.5B model maintained the strongest retrieval performance throughout the benchmark, decreasing from 100.0% at 10 information items to 75.0% at 100 information items. In comparison, the Qwen2.5-0.5B model exhibited the largest decline (90.0% → 40.0%), while the Qwen2.5-3B model maintained intermediate performance (80.0% → 65.0%). These results demonstrate that retrieval capability gradually degrades as contextual information accumulates, although the rate of degradation is not determined solely by model size.

**RQ2. Does successful targeted retrieval necessarily correspond to equally effective global information reconstruction under increasing context loads?**

No. The benchmark demonstrates that retrieval accuracy alone cannot fully represent long-context memory preservation. Although the evaluated models were generally able to retrieve individual facts, their ability to reconstruct the complete information context declined substantially more rapidly as context size increased. This discrepancy was consistently quantified by the proposed Memory Preservation Ratio (MPR), indicating that retrieval capability and reconstruction capability represent complementary dimensions of long-context memory rather than equivalent measures of the same behavior.

**RQ3. How does model scale influence long-context memory preservation when evaluated using complementary retrieval and reconstruction tasks?**

Increasing parameter count did not produce a monotonic improvement in benchmark performance. Across all principal evaluation metrics, including retrieval accuracy, reconstruction performance, and Memory Preservation Ratio, the Qwen2.5-1.5B model consistently outperformed the larger Qwen2.5-3B model. The consistency of this observation across independent evaluation procedures suggests that long-context memory preservation depends on factors beyond parameter count alone and motivates future investigations into the internal mechanisms responsible for this unexpected scaling behavior.

Taken together, these findings demonstrate that evaluating long-context memory requires more than measuring retrieval performance alone. By combining complementary retrieval and reconstruction evaluations with the proposed Memory Preservation Ratio, SRRM provides a more comprehensive characterization of information preservation than existing retrieval-oriented benchmarks. Beyond introducing a reproducible evaluation protocol, this study also demonstrates that complementary memory evaluation can reveal behavioral patterns that remain hidden when only a single retrieval-oriented perspective is considered.

The proposed benchmark is intentionally designed to be lightweight, reproducible, and extensible. Its standardized evaluation pipeline enables future comparisons across different language model families, parameter scales, context lengths, and deployment scenarios with minimal modification. As additional models, datasets, and evaluation metrics become available, SRRM may naturally evolve into a broader benchmark framework capable of supporting increasingly comprehensive investigations of long-context memory behavior.

Ultimately, SRRM is intended not merely as a benchmark for evaluating current language models, but as a reproducible experimental foundation for future memory-oriented research. As language models continue to expand in both parameter scale and context capacity, systematic benchmarks that jointly evaluate retrieval, reconstruction, and information preservation will become increasingly important for understanding how contextual information is retained, integrated, and utilized across increasingly complex long-context applications.

## Acknowledgment

The author would like to express sincere gratitude to John Smith for his valuable discussions, constructive feedback, and insightful suggestions during the early development of this research through the Hugging Face Forum. His comments helped improve the clarity and direction of the proposed benchmark.

The author also gratefully acknowledges the members of the SYNTAX community for fostering a collaborative research environment and providing helpful discussions, encouragement, and technical insights throughout the development of this work.

Finally, the author appreciates the open-source artificial intelligence community for providing publicly accessible language models, software, and research resources that made this study possible.

# Appendix

This appendix provides supplementary materials supporting the reproducibility of the proposed Simple Retrieval-Reconstruction Memory (SRRM) benchmark. It documents the prompt templates, evaluation procedure, and representative examples used throughout the benchmark. Together with the publicly available implementation, these materials enable independent reproduction of the experiments without requiring direct inspection of the source code.

## Appendix A. Benchmark Prompt Templates

The SRRM benchmark evaluates language models through two complementary memory tasks: Targeted Retrieval and Global Information Reconstruction. To ensure experimental consistency, all evaluated models received identical prompt templates, while only the contextual information varied according to the automatically generated synthetic dataset.

Throughout all experiments, the benchmark maintained identical inference settings for every evaluated model. Specifically, all experiments were conducted using a context window of 4096 tokens (num_ctx = 4096) and deterministic decoding with temperature = 0.0. Streaming inference was disabled during the Targeted Retrieval evaluation and enabled only during the Global Information Reconstruction task to facilitate automatic loop detection and real-time reconstruction analysis.

## A.1 Context Construction

Rather than transmitting the entire benchmark context as a single document, SRRM constructs a multi-turn conversation history in which each factual statement is inserted as an individual user message. After every factual statement, the assistant automatically replies with the fixed

```
Understood.
```

acknowledgement:

This procedure simulates a realistic conversational memory scenario while ensuring that every evaluated model receives exactly the same dialogue history.

```
User:
James's hair color is black.

Assistant:
Understood.

User:
Mary went to Tokyo.

Assistant:
Understood.

User:
Kevin's WiFi password is OceanBlue92.

Assistant:
Understood.
```

An example conversation is illustrated below.

The benchmark continues this process until the predefined context size (10, 25, 50, 75, or 100 factual statements) has been fully inserted into the conversation history.

## A.2 Targeted Retrieval Prompt

After constructing the complete conversation history, the benchmark appends a single retrieval question requesting one previously presented fact. The retrieval target is selected automatically from the synthetic dataset, while the question template depends on the attribute category.

The benchmark currently supports the following retrieval templates.

```
What is James's hair color?
Answer short and direct.
```

**Hair Color**

```
Who is Mary's older brother?
Answer only the name.
```

**Family Relationship**

```
Who is Sarah's younger sister?
Answer only the name.
```

Or

```
What did Kevin eat?
Answer only the food.
```

**Food**

```
Where did Daniel go?
Answer only the location.
```

**Location**

```
What is John's WiFi password?
Answer only the exact password.
```

**WiFi Password**

**Other Attributes**

For attributes outside the predefined categories, the benchmark automatically generates a generic

```
Tell me about [Subject]'s [Attribute].
```

retrieval prompt of the form

## A.3 Global Information Reconstruction Prompt

Unlike the retrieval evaluation, the reconstruction task does not request any specific fact. Instead, the

```
You are a strict data logging assistant.

Read all the previous statements carefully.

Do not talk about privacy,
do not ask for missing information,
and do not write any conversational filler.

Your only task is to extract and list every single fact that was mentioned above line by line.

Follow this exact format for each line:

- [Name] [attribute/action] [value]

Example:

- James's hair color is black.
- Mary went to New York.
```

```
Extract all facts now:
```

model receives the following prompt after the complete conversation history has been constructed. The reconstruction prompt intentionally prohibits conversational responses in order to minimize stylistic variation and encourage the model to reproduce only the factual information contained within the previous dialogue history.

## A.4 Automatic Evaluation Procedure

Targeted Retrieval responses are evaluated automatically by comparing the generated answer against the corresponding ground-truth value. The comparison is performed after basic normalization, including lowercasing and removal of punctuation.

Global Information Reconstruction is evaluated using a streaming evaluation pipeline. As each generated line is produced, the benchmark immediately compares the reconstructed statement against the original synthetic dataset. Correctly reconstructed facts are recorded only once, while duplicated information and repeated outputs are tracked separately.

To prevent infinite generation caused by repetitive outputs, SRRM incorporates an automatic loop detection mechanism. The benchmark continuously monitors repeated factual statements during streaming generation. Once the number of repetitions exceeds a dynamic threshold proportional to the number of correctly reconstructed facts, the reconstruction process is terminated automatically. This strategy prevents excessively long generations while preserving all valid information produced prior to termination.

The final reconstruction score is reported as Critical Information Recall (CIR), computed as

$$\text{CIR} = \frac{\text{Number of Correctly Reconstructed Facts}}{\text{Total Number of Facts}}$$

## Appendix B. Synthetic Dataset Examples

This appendix presents representative examples of the synthetic dataset used throughout the experiments. All contextual information was generated automatically using the SRRM dataset generator under controlled conditions. Consequently, every evaluated language model received identical factual information, ensuring fair comparison across all benchmark configurations.

### B.1 Example Context

A representative subset of generated contextual information is shown below.

[1] James's hair color is black.
[2] Mary went to Tokyo.
[3] Kevin ate pizza.
[4] John's WiFi password is OceanBlue92.
[5] Sarah's younger sister is Emily.
[6] Daniel went to Paris.
[7] Michael ate sushi.
[8] Anna's hair color is blonde.
[9] Olivia's older brother is Ryan.
[10] Lucas's WiFi password is GreenApple88.

### B.2 Example Targeted Retrieval Evaluation

The following example presents a representative Targeted Retrieval evaluation performed by the SRRM benchmark. During each benchmark run, the synthetic dataset is first generated and sequentially inserted into the conversation history. After the contextual information has been established, the benchmark automatically generates retrieval questions for every previously presented fact in reverse chronological order while maintaining an identical conversational context for each evaluation.

**Generated Context**

[1] Steven's hair color is grey.
[2] Hannah has a younger sister named Jose.
[3] Melissa went to Berlin.
[4] Gloria's WiFi password is h2lcSPcfE.
[5] Raymond's hair color is blonde.
[6] Jason has a older brother named Alice.
[7] Shirley went to Sydney.
[8] Christian's WiFi password is dWO4DDOYf.
[9] Jason's hair color is black.
[10] Lauren has a younger brother named Karen.

What is Christian's WiFi password?
Answer only the exact password.

**Example Retrieval Query**

dWO4DDOYf

**Ground Truth**

dWO4DDOYf

**Model Response**

Result: CORRECT

**Evaluation Result**

During the benchmark, every factual statement is evaluated independently using the corresponding retrieval template. The benchmark records the predicted answer, correctness, input tokens, output tokens, inference latency, peak RAM usage, and peak VRAM usage for each retrieval query before calculating the final retrieval accuracy.

## B.3 Example Global Information Reconstruction Evaluation

After completing the Targeted Retrieval evaluation, the benchmark performs a Global Information Reconstruction task using the same conversational context. Instead of requesting a specific fact, the language model is instructed to reconstruct every previously presented statement from memory.

```
You are a strict data logging assistant.

Read all the previous statements carefully.

Do not talk about privacy,
do not ask for missing information,
and do not write any conversational filler.

Your only task is to extract and list every single fact that was mentioned above line by line.

Follow this exact format for each line:

- [Name] [attribute/action] [value]

Example:

- James's hair color is black.
- Mary went to New York.

Extract all facts now:
```

**Reconstruction Prompt**

```
- Steven's hair color is grey.
- Hannah has a younger sister named Jose.
- Melissa went to Berlin.
- Gloria's WiFi password is h2lcSPcfE.
- Raymond's hair color is blonde.
- Jason has a older brother named Alice.
- Shirley went to Sydney.
- Christian's WiFi password is dWO4DDOYf.
```

**Example Model Output**

```
Correctly reconstructed facts : 8
Total facts                   : 10
Critical Information Recall   : 0.80 (80%)
Loop detected                 : No
```

**Automatic Evaluation**

During reconstruction, every generated line is immediately compared against the original synthetic dataset using the SRRM evaluation pipeline. Correct facts are counted only once, duplicate facts are ignored for recall calculation, and repeated outputs are continuously monitored by the loop detection mechanism. Once the reconstruction process terminates, the benchmark computes the final Critical

Information Recall (CIR) score together with the associated computational statistics, including inference time, token usage, peak RAM usage, and peak VRAM usage.